\documentclass[lettersize,journal]{IEEEtran}
\usepackage{amsmath,amsfonts}
\usepackage[noend]{algpseudocode}
\usepackage[ruled,vlined,linesnumbered]{algorithm2e}
\algrenewcommand{\Return}{\State \textbf{return}\ }
\algnewcommand{\funccall}[1]{\textit{#1}}
\usepackage{array}
\usepackage[caption=false,font=normalsize,labelfont=sf,textfont=sf]{subfig}
\usepackage{textcomp}
\usepackage{stfloats}
\usepackage{url}
\usepackage{verbatim}
\usepackage{graphicx}
\usepackage{cite}
\begin{document}

\title{From Interpretable Filters to Predictions of Convolutional Neural Networks with Explainable Artificial Intelligence}

\author{Shagufta Henna,~\IEEEmembership{Senior Member,~IEEE,}
Juan Miguel Lopez Alcaraz,
\thanks{Shagufta Henna and Juan Miguel Lopez Alcaraz are with the Department of Computing, Atlantic Technological University, Donegal, Ireland}
\thanks{Manuscript received April 19, 2021; revised August 16, 2021.}}

\markboth{IEEE Transactions on Artificial Intelligence,~Vol.~XX, No.~X, XXX~2022}%
{Shell \MakeLowercase{\textit{et al.}}: A Sample Article Using IEEEtran.cls for IEEE Journals}

\IEEEpubid{0000--0000/00\$00.00~\copyright~2021 IEEE}

\maketitle

\begin{abstract}
Convolutional neural networks (CNN) are known for their excellent feature extraction capabilities to enable the learning of models from data, yet are used as black boxes. An interpretation of the convolutional filtres and associated features can help to establish an understanding of CNN to distinguish various classes. In this work, we focus on the explainability of a CNN model called as $cnn_{explain}$
 that is used for Covid-19 and non-Covid-19 classification with a focus on the interpretability of features by the convolutional filters, and how these features contribute to classification. Specifically, we have used various explainable artificial intelligence (XAI) methods, such as visualizations, SmoothGrad, Grad-CAM, and LIME to provide interpretation of convolutional filtres, and relevant features, and their role in classification. We have analyzed the explanation of these methods for Covid-19 detection using dry cough spectrograms. Explanation results obtained from the LIME,  SmoothGrad, and Grad-CAM highlight important features of different spectrograms and their relevance to classification.

\end{abstract}


\section{Introduction}
\IEEEPARstart{A}{rtificial}  intelligence (AI)  has several applications in various industries to improve goods, methods, and research.  Recent years have demonstrated significant success in the AI subfield known as deep learning (DL)/neural networks due to its good performance for multidimensional data. The deep learning model consists of multiple processing layers, to model complex nonlinear input-output relationships, and the ability to perform pattern recognition and feature extraction from low-level input data. 
Although, these deep learning approaches, can capture non-linear relationships in input data, however, often comes at the cost of limited interpretability. DL models are notorious to elude immediate interpretability by humans, thereby limiting their applications in certain situations, such as medical diagnosis. 
Explainable artificial intelligence (XAI) aims to explain and interpret the internals of machine learning or deep learning models \cite{Das2020}. The role of XAI in medical diagnosis plays a critical role, due to clinicians who want to consider AI-driven decisions as a second opinion. The explainability of “how” the deep learning model makes a prediction may seem a global explanation of the network, however,  the explainability of the “why” is more inclined toward the contribution of features towards classification.

Undoubtedly, XAI provides a better environment for deep learning applications in many ways, and it comes with limitations and challenges. As an example, when it comes to the raw explainable results based on visualization, there are two main concerns as reviewed by \cite{Das2020}. Firstly, there is human attention incapacity to deduce XAI decision-making explanation maps. Secondly, appropriate explainable AI techniques are unavailable to quantitatively provide better interpretability and explanation to augment visualizations. 

The challenges that XAI present in healthcare have recently been investigated by various researchers, specifically for the clinical decision support system (CDSS) \cite{Antoniadi2021}. A work by \cite{Cai2019} found that pathologists who had local model interpretations before the model adoption require more insights on model  characteristics for complete trustworthiness. Further, authors in \cite{Liao2020} show that identifying the rationale for explainability aids in the selection of XAI approaches that can fill gaps when creating user experiences. Another work by \cite{Tan2008} suggest that to obtain user acceptability, a CDSS should have high tractability, which necessitates human reasoning, stepwise inference, explanation capability, and user-familiar terminology.  Overall, the explainability for CDSS demands easily understandable terminologies and an explanation of the neural network architecture.

The trend of XAI for medical applications is relatively new, however, there are some efforts in this domain to explain the role of deep neural networks. \cite{Lee2020} implemented a gradient-weighted class activation map (Grad-CAM) algorithm for Covid-19 features detection using X-ray images. Another similar work by \cite{Montebello2021} implemented a layer-wise relevance propagation (LRP) algorithm to highlight breast cancer features in mammogram images. \cite{Li2021} proposed the Grad-CAM algorithm and its derivatives for the explainability of traffic accident anticipation in automated driving systems.  Another work in \cite{Kim2020} retrieved important job skills from a job classification task with local interpretable model-agnostic explanations (LIME) algorithm.  Authors in \cite{Nigri2020} proposed an explainable algorithm that analyzes the performance of various explainable algorithms by removing noise for disease detection. \cite{Ahn2020} implemented an explainable algorithm based on dimensionality reduction for traffic classification. \cite{Ren2020} presented an explainable method based on a hybrid neural network for acoustic scene classification.

To our best knowledge, all the recent works on the explainability of neural networks focus on model internals.  There is no work,  specifically convolutional neural networks (CNN) to explain the CNN filters to interpret features and their role in Covid-19 classification.

This research focuses on the explainability of CNN, a popular neural network for image classfication with a particular focus on its filtres. Further, it also investigates various explainability approaches for the local explanation of features interpreted by CNN filtres and the decisions made by the classification layer. The  major contributions are listed as follows:

\begin{itemize}
\item  We evaluate the accuracy of CNN for Covid-19 detection using dry cough to show its suitability for explainability. From an application point of view, this is different from \cite{Lee2020} and \cite{Alshazly2021} because their approaches detect Covid-19 based on X-ray and CT images respectively. On the other hand, our work focuses on cough sounds that can support a faster and cheaper diagnosis.

\item  We presented features visualization of CNN to show a positive correlation between the complexity of patterns learned by the network. The proposed convolutional filters visualization is the most relevant pattern in a deep neural network. On the contrary, this work puts an effort to analyze the patterns learned by the filters of each hidden layer of CNN.

\item  We showed that the visualization of CNN layers helps to understand a clear and distinct characteristic between a positive and a negative class to help to interpret the model predictions. Our work focuses on features that the CNN related to each class using the process of activation maximization. This is different from the work of \cite{Gong2020} which uses learned patterns to distinguish the classes.

\item We evaluated various XAI methods such as Smooth-Grad, Grad-CAM, and LIME to explain the CNN filtres and classification when applied for Covid-19 detection using dry cough. 
This differs from the existing works that focus on the application of saliency maps to explain model internals, and segmentation by occlusion to evaluate the feature importance using inputs and outputs. 
\end{itemize}

\section{Convolutional Neural Network for Covid-19 Detection} \label{model}

\begin{figure}[!t]
\centering
\includegraphics[width=3.5in]{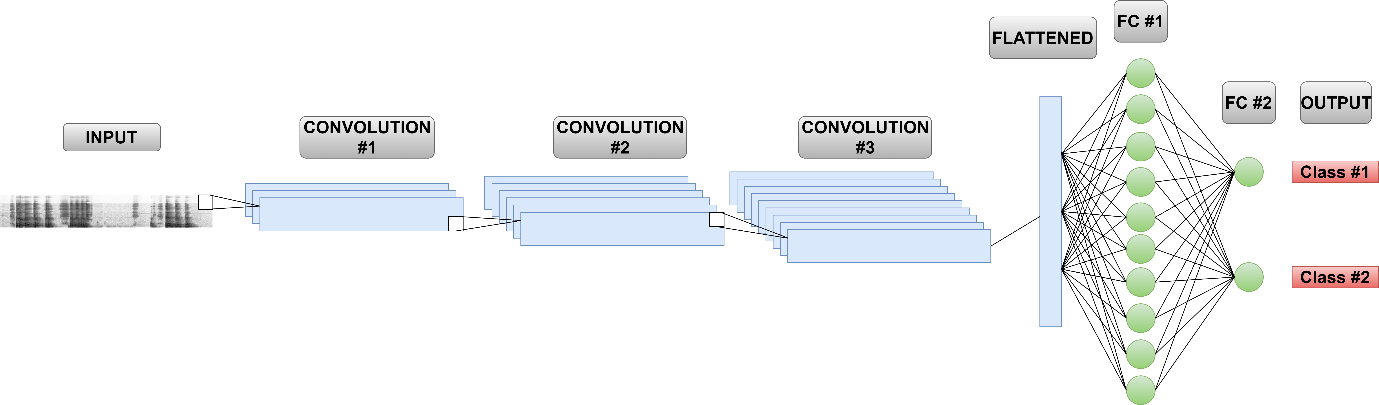}
\caption{Convolutional neural network for Covid-19 detection.}
\label{cnn}
\end{figure}

This section presents the CNN model used for Covid-19 classification using dry cough spectrograms. The model internals as shown in Figure  \ref{cnn} that are later explained using various XAI approaches in Section \ref{xai1}.

Figure \ref{cnn}  shows that the CNN takes a spectrogram as input and consists of three convolutional layers with the primary function to extract features from the pixeled images. This is followed by a flattening process converting the two-dimensional image to a single-dimensional matrix suitable for the fully connected layer. The fully connected layer enables learning based on non-linear combinations of extracted features and its output is passed to another fully connected layer that outputs the probabilities of Covid-19 and non-Covid-19 classes.

Algorithm \ref{alg:cnn} shows the steps of CNN used for Covid-19 detection.  The algorithm performs classification as Covid-19 or non-Covid-19 using the spectrogram input. It computes the loss using the predicted label $\hat y$ and true label $y$ that is later backpropagated to adjust the weights and bias using an optimizer. The basic steps given in Algorithm \ref{alg:cnn} are explained in the sections below.

\begin{algorithm} \label{alg:cnn}

  \SetAlgoLined
\KwData{ $Spectrogram$ $X$, $w$,$b$}
 \KwResult{Model $cnn_{explain}$}  
   \For {each epoch}{        
 \For {each batch x $\in X$ and c}{
$ z_t^{(1)}=\mathbb{R}\left(w^{(1)}x+b^{(1)}\right)$ \\
$z_t^{(1)}=MaxPool\left(z_t^{(1)}\right)$ \\
$z_t^{(1)}=DropOut\left(z_t^{(1)}\right)$ \\
$z_t^{(2)}=\mathbb{R}\left(w^{(2)}z_t^{(2-1)}+b^{(2)}\right)$ \\
$z_t^{(2)}=MaxPool\left(z_t^{(2)}\right)$ \\
$z_t^{(2)}=DropOut\left(z_t^{(2)}\right)$ \\
$z_t^{(3)}=\mathbb{R}\left(w^{(3)}z_t^{(3-1)}+b^{(3)}\right)$ \\
$z_t^{(3)}=MaxPool\left(z_t^{(3)}\right)$ \\
$z_t^{(3)}=DropOut\left(z_t^{(3)}\right)$ \\
$z_t^{(f)}=Flatten\left(z_t^{(3)}\right)$ \\
$z_t^{(4)}=\mathbb{R}\left(w^{(4)}z_t^{(f)}{+b}^{(4)}\right)$ \\
$z_t^{(4)}=DropOut\left(z_t^{(4)}\right)$ \\
$z_t^{(5)}=(\hat y)=\sigma(w^{(4)}z_{t}^{(4)}+b^{(4)})$\\
$cnn_{explain} \leftarrow z_t^{(5)}$ \\
$\hat y \leftarrow cnn_{explain}.predict()$ \\
$cnn_{explain}loss=y-(\hat y)$ \Comment{calculates loss} \\
backpropagation($cnn_{explain}$) \Comment{backpropagate loss} \\
Optimize($cnn_{explain}$)  \Comment{ minimize loss } \\

	  	             }                        
                 } 
                                            
\caption{CNN model for Covid-19 detection} 
\end{algorithm} 

\subsection{Dry Cough Audio to Spectrograms}
\begin{figure}[!t]
\centering
\includegraphics[width=3.5in]{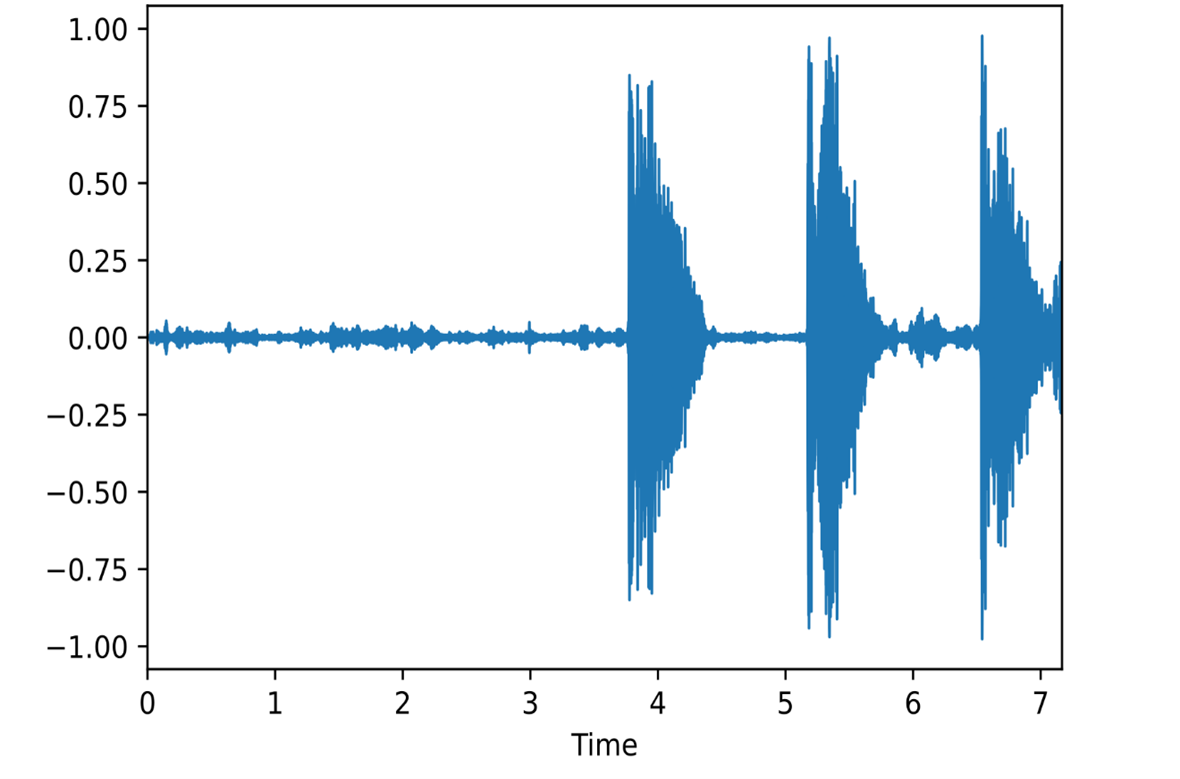}
\caption{Wave plot of dry cough audio.}
\label{wplot}
\end{figure}

The neural network in Algorithm \ref{alg:cnn} takes spectrograms as an input. These spectrograms are the result of conversion from audio to images, representing the meaningful audio information to train a convolutional neural network. Figure \ref{wplot} shows the wave plot of the dry cough audio. The y- axis represents the amplitude of the cough sound, whereas, the x-axis shows the time of the audio using a sampling rate of  44,100 hertz. The wave plot in the Figure shows approximately 308,700 data points for 7 seconds that reflects the information contained in the audio.

\begin{figure}[!t]
\centering
\includegraphics[width=3.5in]{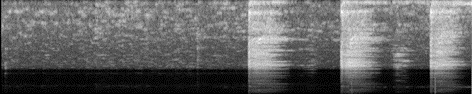}
\caption{Scaled Mel spectrogram.}
\label{scalspectro}
\end{figure}

Figure \ref{scalspectro} shows the scaled Mel spectrogram generated using two steps: magnitude spectrogram generation, and mapping from spectrogram to Mel scale. The magnitude spectrogram is computed by applying the short-time Fourier transform (STFT) \cite{Boashash2015}.

\begin{equation}
Y \left(m,k\right)=\ \left|S(m,k)\right|^2
\end{equation} \label{eq1:spect}

SFTF generates a matrix of magnitudes that are later squared to obtain an image representation of spectrograms as given in Equation \ref{eq1:spect}. In the Equation, $m$ denotes the signal and $k$ as the window of the spectrogram computation. Later, the frequencies of magnitude spectrogram are mapped to the Mel scale as given in Equation \ref{eq1:freq} \cite{Cai2011}. The scaled Mel-spectrogram in Figure \ref{scalspectro} shows low frequencies after amplification and high frequencies using compression with the help of the Minmax scaling process to fit the values into the 8-bit range, thereby retaining significant information in a smaller size image.

\begin{equation}  
m=2595{log}_{10}\left(1+\frac{f}{700}\ \right)
\end{equation} \label{eq1:freq}

\begin{figure}[!t]
\centering
\includegraphics[width=3.5in]{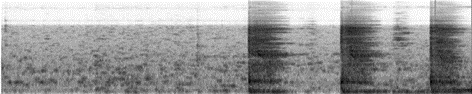}
\caption{Mel spectrogram colour exchange and image flipping.}
\label{flipspectro}
\end{figure}
Figure \ref{flipspectro} shows the scaled Mel spectrogram using inversion of color intensity, where black areas represent high and white as low energies, respectively. This results in a final spectrogram, where the flipping of the image moves the low frequencies at the bottom, making it ready as the input to the CNN.

\subsection{CNN Initialization}
Algorithm \ref{alg:cnn} takes the weights $w$ and biases $b$ to train all the layers of the CNN that are later adjusted to minimize the training cost or loss. The weights in convolutional layers are interpreted as kernels, i.e., a weight matrix multiplied with input to extract relevant features. For 2D convolutions, kernels, as used in the Algorithm  \ref{alg:cnn}, are also called filtres. The bias shifts the activation function by adding a constant value to it.
The weights of the neural network are initialized using the Glorot uniform based on the size of the neural network as given in Equation \ref{eq1:weight} \cite{Glassner2021}. The Equation randomly initializes the weights of the neural network using the number of input and hidden units in the neural network. Using the $4$ input and $5$ hidden units, the Equation calculates the random weights in a range of -0.82 to +0.82. The bias is initialized with a value of $0$.

\begin{equation}
{GU}_{limits}=\ \sqrt{\frac{6}{N_{in}+N_{out}}} 
\end{equation} \label{eq1:weight}

\subsection{Feature Extraction using Convolutional Filtres}
Figure  \ref{flipspectro} illustrates the basic process of applying a kernel or filter to the normalized spectrogram. It shows the complete feature map after passing the filter through all the possible positions on the Mel-Spectrogram. The filter represents the weight with the values initialized using the Glorot uniform function given in Equation \ref{eq1:weight}. Wee have considered a filter size of $3 \times 3$. However, the size can be adjusted based on the feature retrieval specific to the area of a Mel-Spectrogram.
This filter is later multiplied with the dry cough spectrogram, followed by a dot product with added bias later to generate a value \cite{Brownlee2019}. Finally, the resultant values are added to a feature map. The same filter convolves from left to right and up to down with the Mel-Spectrogram adding more values to the feature map. The values in the feature map correspond to Mel-Spectrogram pixels while maintaining lass-specific pixel correlation.
 
\begin{figure}[!t]
\centering
\includegraphics[width=3.5in]{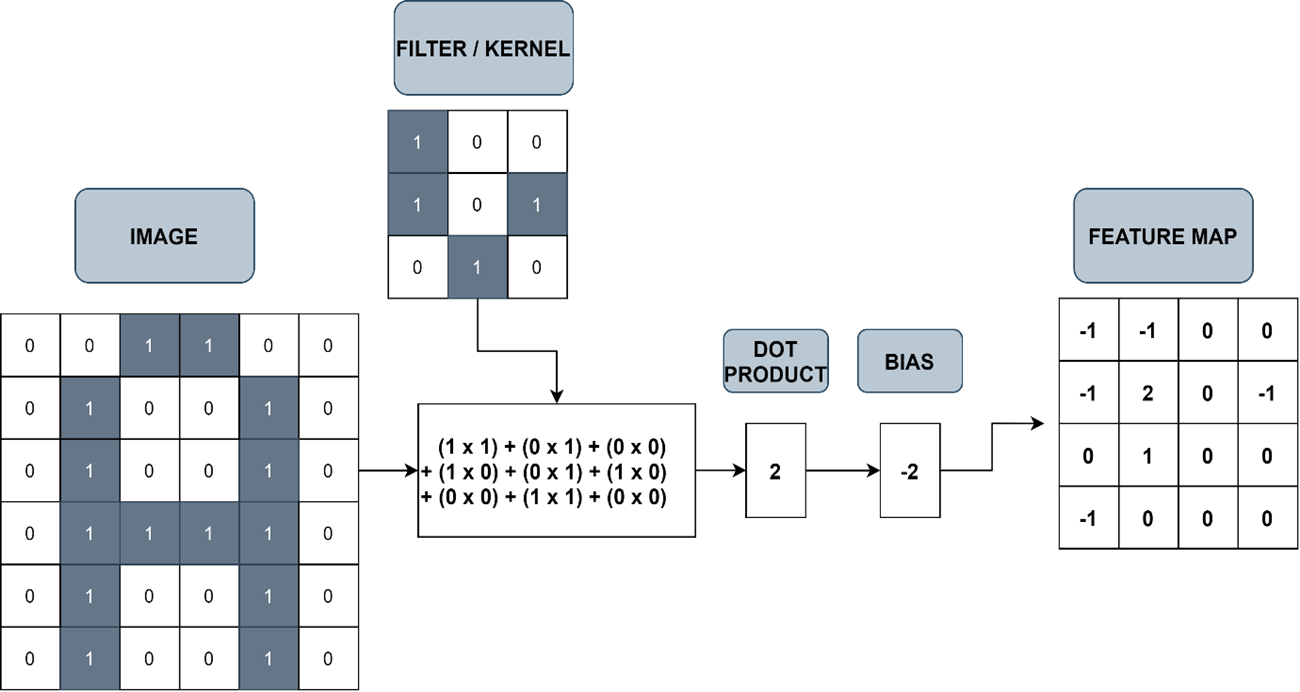}
\caption{Complete feature map generated by the convolutional filters.}
\label{flipspectro}
\end{figure}
\subsection{Activation and Max Pooling}
Equation \ref{eq1:relu} shows the activation function for the first three convolutional and the first fully connected layer of the CNN. The activation function is based on the Rectified Linear Unit (ReLu) activation with values between 0 and the input value (Nair and Hinton, 2010).
\begin{equation}
\mathbb{R}\left(z\right) =max(0,z) 
\end{equation} \label{eq1:relu}
After the activation function, Max pooling is applied to the feature map. Max pooling is considered another new filter, however, it takes only the maximum value from each feature map region, constituting a reduced size feature map. In Algorithm \ref{alg:cnn},  we have considered a pool size of $2 \times 2$, although its size can be adjusted based on the input and convolutional filters.

\subsection{Dropout and Flattening}
Once Max pooling is complete, dropout is applied for the removal of input-output connections to regularize the network. In the neural network for Covid-19 classification, the dropout process drops approximately 25\% of the features and is applied $4$ times in Algorithm \ref{alg:cnn}.
Once the dropout process is complete, we consider the flattening process for the CNN that converts the 2-dimensional feature map to a 1-dimensional feature space to feed to the fully connected layer as given in Algorithm \ref{alg:cnn} \cite{Bhateja2019}. 

The fully connected layer also known as the dense layer connects each unit of the last layer with all the units of the same layer. Our model used for explainability as given in Algorithm \ref{alg:cnn} uses two dense layers, each with its weights and bias. 

\subsection{Covid-19 Classification and Training}
This section illustrates the Covid-19 and non-Covid-19 classification using the dry cough spectrogram using the CNN  Algorithm \ref{alg:cnn}. This classification is explained later using various explainable AI approaches. Algorithm \ref{alg:cnn} shows the use of the SoftMax activation $\sigma(z)$ function as given in Equation \ref{eq1:sigma} based on the features extracted by the convolutional filters. SoftMax function assigns a confidence score to each class by applying an exponential function.

\begin{equation}
{\sigma(z)}_i=\ \frac{e^{zi}}{\sum_{j=1}^{k}e^{zj}}
\end{equation} \label{eq1:sigma}

Equation \ref{eq1:softmax} shows how SoftMax function calculates the probability $\mathbb{P}$ of each class $y$ at given timestep $t$ from the $k$ classes, i.e., Covid-19 and non-Covid-19 from the input features $f$. The function results in a predicted class called $\hat y$.

\begin{equation}
\hat y_{t,k} \equiv \mathbb{P}\left(y_t=k|x\right)= \frac{\exp(w_k^{(5)}h_t^{(4)}+b_k^{(5)})}{\sum_{j}{exp(w_j^{(5)}h_t^{(4)}+b_j^{(5)})}}  
\end{equation} \label{eq1:softmax}

Line 16-18 illustrates the loss computation, i.e., the difference between true class label $y$ and predicted class label $\hat y$, that is backpropagated from the output layer to the input layer to adjust the weights of each layer using the Adam optimizer. Adam optimizer is based on adaptive moment estimation. It incorporates anticipated values $E$ of previous gradients as given in Equation \ref{eq1:adam}.  Adam optimizer function calculates momentum using various steps consisting of distinct parameters, thereby leading to faster convergence.

\begin{equation}
\theta_{t+1,i}= \theta_{t,i}- \frac{\eta}{\sqrt{E[G}_{t,ii}+\epsilon]}  \cdot{E[g}_{t,i}]
\end{equation} \label{eq1:adam}

\section{Explainability of CNN for Covid-19 Detection} \label{xai1}
This section discusses the explainable AI approaches to help interpret the deep learning model given in Algorithm \ref{alg:cnn} and explained in Section \ref{model}.  This section starts with the visualization of features to help interpret the role of these features in the Covid-19 classification.  After feature visualizations, we have used SmoothGrad Algorithm to highlight the importance of features of a spectrogram to Covid-19 predictions. This is followed by the GradCAM to provide visualization of the spectrogram to  indicate the important regions in a feature map in the last convolutional layer as given in Algorithm \ref{alg:cnn}. Finally, explainable AI approach LIME provides the feature importance of a spectrogram using segments.

\subsection{Feature Visualization by Optimization} \label{exmethod}
We have used two approaches of visualization to explain the features and the learning of convolutional neural network given in Algorithm \ref{alg:cnn}. The first approach as given in Algorithm \ref{alg:feature} helps to interpret convolutional filters using visualization, whereas the second one given in Algorithm \ref{alg:dense} provides an interpretation of dense layers in Algorithm \ref{alg:cnn}. These algorithms involve the use of a loss function to maximize the activation of desired neurons within the selected layer.

Algorithm \ref{alg:feature} helps to understand the first convolutional layer filter for simple textures, and the second from Algorithm \ref{alg:dense} for complicated textures and patterns from the dry cough spectrogram. The last convolutional filter in Algorithm \ref{alg:cnn} learns objects or fragments of objects from the spectrogram. In the case of dry cough spectrograms, these objects are various frequencies activated according to the dry cough audio, thereby making its interpretation significantly challenging.

\begin{algorithm} \label{alg:feature}

  \SetAlgoLined
\KwData{ $cnn_{explain}$ from Algorithm \ref{alg:cnn}}
 \KwResult{Filter $A_{desired}$}  
 $model_{filters} \leftarrow \emptyset$ \\
   \For {each $layer_{i}$ in $cnn_{explain}$}{        
\If {$layer_{i}==covolutional$} {
$layer_{i}^{conv} \leftarrow layer_{i}$
         }
 $layer_{i}^{conv}.set\_activation(linear)$ \Comment{set the model activation as linear } 
  $model_{filters}.inputs \leftarrow cnn_{explain}.inputs$ 
    $model_{filters}.outputs \leftarrow layer_{i}^{conv}.outputs$ \Comment{set the model activation as linear }
   }
  \For {each $filter_{i}$ in $model_{filters}$}{ 
    \If {$filter_{i}$ is desired}{
    $A_{desired} \leftarrow filter_{i}$ \\
   return  $A_{desired}$  \Comment{return the desired filter for visualization } 
       }}

\caption{Feature visualization of convolutional filters.} 
\end{algorithm} 

\begin{algorithm} \label{alg:dense}
  \SetAlgoLined
\KwData{ $cnn_{explain}$ from Algorithm \ref{alg:cnn}}
 \KwResult{$model_{dense}$}  
   \For {each $layer_{i}$ in $cnn_{explain}$}{        
\If {$layer_{i}==dense$} {
$layer_{i}^{dense} \leftarrow layer_{i}$
         }
 $layer_{i}^{dense}.set\_activation(linear)$ \Comment{set the model activation as linear } \\
  $model_{dense}.param \leftarrow cnn_{explain}.param$ \\
    $model_{dense}.outputs \leftarrow layer_{i}^{dense}.outputs$ \Comment{set the dense model output as outputs of the dense layer }
   }
  \For {each $unit_{k}$ in $model_{dense}$}{ 
     $img_{k} \leftarrow creatImage(model_{dense}.inputSample)$ \\
  $model_{dense} \leftarrow maxActivation(img_{k},unit_{k}$)  \Comment{maximize activation of Image in each unit of dense layer} 
       }

\caption{Feature visualization of dense layer.} 
\end{algorithm} 

Algorithm \ref{alg:feature} presents the steps required for the visualization of convolutional filter. The first step extracts the required convolutional layer from the CNN model. Later, it enables the linear activation function to take the raw input. This is followed by the creation of a new model with the same inputs as the first model and the output of the last layer. Slicing the model into the desired number of convolutional layer transforms it to a new model  that does not change the first model by collecting its outputs. Lines 9 to 12 iterates over desired filters for visualization returning an image  to distinguish the Covid-19 or non-Covid-19 classes based on specific neural network level patterns. These patterns along with the filters and parameters are learned during the first model training.

Algorithm \ref{alg:dense} lists the steps used for the visualization of the last dense layer using the SoftMax classification. The first step in the algorithm extracts the desired layer and changes the SoftMax as an activation function for the analysis. This change  brings desirable linearity to the model outputting the probabilities  of the Covid-19/non-Covid-19 classes with a sum of 1. The second sets the layer's activation function to linear.  Similar to Algorithm \ref{alg:dense}, the third creates a new model based on the specification of the first model. This is to prevent the modification to the existing model while explicitly revealing the CNN model internals at this stage.  This is followed by an iterative process to apply the activation maximization function that generates a random and optimized image for Covid-19/non-Covid-19 as predicted by the Algorithm \ref{alg:cnn}.

\subsection{Smooth Gradient (SmoothGrad) for cnn$_{explain}$ Model Interpretation}
This section presents SmoothGrad to interpret cnn$_{explain}$ in Algorithm \ref{alg:cnn}. The method focuses on the explanation of the feature importance of input dry cough Mel-spectrogram image as given in Figure \ref{scalspectro} and outputs a sensitive map. Specifically, the SmoothGrad aims to maximize the class score as given in Equations \ref{eq:smooth1} and \ref{eq:smooth2}.

\begin{equation} \label{eq:smooth1}
class\left(x\right)=argmax_{c \in C} S_{c}(x)\\ 
\end{equation}

\begin{equation} \label{eq:smooth2}
M_{c}\left(x\right)=\frac{\partial S_c(x)}{\partial x}  
\end{equation}

\begin{algorithm} \label{alg:smoothgrad}
  \SetAlgoLined
\KwData{ Image $x$, noise spread level $N(0,\sigma^{2})$, number of samples $n$,$cnn_{explain}$ model }
 \KwResult{ SmoothGrad sensitive map ${M}^{'}_{c}$}
$ totalImages \leftarrow x.{n}$
  
   \For {each image $x$ in totalImages}{        
     $x \leftarrow x+N(0,\sigma^{2})$ \\
     $S_{c}(x) = cnn_{explain}.predict(x)$ \\
     $M_{c}(x)=\ \frac{\partial S_c(x)}{\partial x}$ \\
     ${M}^{'}_{c} \leftarrow \frac{\sum_{i=1}^{n}M_{ci}(x)}{n}$
     \Comment{sensitive map smoothed with noise base } 
       }

\caption{SmoothGrad.} 
\end{algorithm}

For the given input features $x$ extracted from the dry cough Mel-spectrogram, Activation function in Equation \ref{eq:smooth1} selects a class $c$, i.e, covid-19/non-covid-19 from the set of classes $C \in \{$Covid-19, non-Covid-19$\}$ to maximize the class score. Equation \ref{eq:smooth2} computes a sensitive map $M_{c}\left(x\right)$ for the input Mel-spectrogram where  $\partial S_{c}$ represents the derivative gradient of the class score function for input $\partial x$. This measures any change in classification score concerning the change in each feature, i.e., pixels of the Mel-spectrogram. This map, however, also includes noise or random pixels in Mel-spectrogram. This is attributed to the local variations in score function  $S_{c}$ during the backpropagation process

Equation \ref{eq:smooth3} selects the saliency map ${M^\prime}_{C}\left(x\right) $ of a Mel-spectrogram estimating the local average of the gradient values with a stochastic approximation.  ${M^\prime}_{C}\left(x\right) $ is computed by  sampling similar images by adding random noise with a gaussian distribution with a mean of 0 and standard deviation to set $N\left(0,\ \sigma^2\right)$, then averaging the resultant sensitive maps for each Mel-spectrograme for $n$ samples.

\begin{equation} \label{eq:smooth3}
{M^\prime}_C\left(x\right)=\ \frac{1}{n}\sum_{1}^{n}{M_c(x+N\left(0,\ \sigma^2\right))}
\end{equation}

Algorithm \ref{alg:smoothgrad} shows steps for sensitive map calculation. Lines 2-6 illustrate the iterative steps to apply noise using a gaussian distribution with a mean of 0 and a standard deviation$ N\left(0,\ \sigma^2\right)$ to each Mel-spectrogram. This augmented Mel-spectrogram $x$ is then classified by the cnn$_{explain}$ model returning a class score. The application of noise during this process improves sensitive map called as stochastic resonance process. This process replicates the Mel-spectrogram with random noise that strengthens the pixel patterns between the Mel-spectrogram samples without amplifying the random noise.  Line 5 calculates the sensitive map with the partial derivatives of the class score ${M^\prime}_C\left(x\right)$ for the augmented Mel-spectrogram $x$. Finally, line 6 returns the sensitive map smoothed with noise base on the average of sensitive maps $\frac{\sum_{i=1}^{n}M_{ci}(x)}{n}$ for each augmented Mel-spectrogram. 
Later, during the performance evaluation, Algorithm \ref{alg:smoothgrad} provides meaningful insights for Covid-19/ non-Covid-19 class for the input Mel-spectrogram.

\subsection{Gradient-weighted Class Activation Mapping (Grad-CAM)}
 Similar to SmoothGrad, Grad-CAM outputs a sensitivity map for a Mel-spectrogram image. However, in contrast to SmoothGrad , it emphasizes the areas of importance by the feature maps in the last convolutional layer for the Covid-19/non-Covid-19 classification. Extracting the importance of pixels in a Mel-spectrogram by the last layer/feature map leads to the collection of both the Covid-19 and non-Covid-19 classes that can be predicted. In our work, we focus on a single class to conduct an analysis of class differentiation using the activation of each sample area. 

To collect the importance of each pixel in a spectrogram for image classification, derivatives by backpropagation are required. This process can lead to confusion in the feature's importance collection during training. However, in this work, we have considered the backpropagation at an inference phase. This is given in Algorithm \ref{alg:gradcam}. It computes the global and averaged pooled gradients $\alpha_k^cof $ of each feature map $A^k$ passed through the ReLU.

\begin{algorithm} \label{alg:gradcam}
  \SetAlgoLined

\KwData{ Image $x$, class $c$,$cnn_{explain}$ model }
 \KwResult{ Sensitive map $GC^{c}_{+}$}
 forward $x$ in $cnn_{explain}$ up to $h_t^{-1}$ \\
 $y^{c} \leftarrow collect(c)$ 
  \Comment{Collect raw score of desired class $c$. } \\
   $y^{!c}=0$ 
 \Comment{set other classes to 0.  } \\
  
   \For {each $A \in h_t^{-1}$}{ 
   
              $\frac{dy^{c}}{dA^{k}}$ 
       \Comment{Backpropagate $c$ score in feature maps of the last layer $A^{k}$. } 
       }
          
    \For {each $A \in h_t^{-1}$}{ 
              $\alpha_k^c=\ \frac{1}{Z}\sum i_{width}\sum j_{height}  $ 
       \Comment{Global average pooling of gradient for each feature map. } 
       }
          
$GC^{C}=\alpha_1A^1+\ \alpha_2A^2\ldots\ \sum_K\alpha_k^cA^k$
\Comment{Weighted sum of feature maps. } \\
  $GC^{c}_{+}=ReLu(max,{GC}^{c})$
  \Comment{ReLu function. } 
                                                            
\caption{Grad-CAM} 
\end{algorithm}  

Algorithm \ref{alg:gradcam} presents all the steps by the GradCAM to compute the sensitive map.
Line 1 passes the Mel-spectrogram using the forward pass in the last convolutional layer of the  $cnn_{explain}$ model.
Line 2-3 collects the score of the desired class $c$, i.e., Covid-19 or non-Covid-19, and sets the score of the other class to $0$. Lines 4-6 refer to the backpropagation of the class score for the Mel-spectrogram in the convolutional filters to calculate the gradients, representing the feature importance of the  Mel-spectrogram as $ A^k$.  Lines 7-9 iteratively perform global average pooling of gradients of each feature map, returning the weights of each feature map. Line 10 computes the weighted sum of the feature maps with their respective weights, called as the sensitive map.

 The first step is to send an image input in a forward pass in the model up to the last convolutional layer. The second step is to collect the score of the desired class and set the other classes' scores to zero in step 3. Step number four refers to the backpropagation of the class score obtained concerning the input image but only in the convolutional filters of the convolutional last layer to obtain the gradients, representing the feature importance of the image. Lines 7-9 iteratively perform global average pooling of gradients of each feature map, returning the weights of each feature map. Line 10 computes the weighted sum of the feature maps with their respective weights, called the sensitive map.  In the later step, negative values in the sensitive map are eliminated using the ReLu function, thereby highlighting only the positively correlated features in the Mel-spectrogram.

W have resized the sensitive map to facilitate its visualization and comparison for the corresponding Mel-spectrogram. This is needed as the convolutional filters of the last layer are smaller than the original Mel-spectrogram input because of the image transformation. Algorithm \ref{alg:gradcam}results in map  $GC^{c}_{+}$ highlighting the most important areas of a Mel-spectrogram by each feature map $A$ corresponding to the last convolutional layer $h_t^{-1}$ contributing significantly to the prediction covid-19 or non-covid-19 class.

\subsection{Local Interpretable Model-Agnostic Explanations (LIME)}

The fourth method, we have considered for the explainability of cnn$_{expain}$  is called the LIME algorithm. LIME considers image classification with a focus on specific areas of an image contributing to the Covid-19/non-Covid-19 class. In contrast to SmoothGrad and GradCAM that average sensitive maps and compute a sensitive map from the last filter, the LIME algorithm aims to identify the areas of importance by occlusion. 
LIME uses an interpretable model to interpret a model by approximating the black-box cnn$_{explain}$.  This interpretable model supports straightforward computations to understand than the algorithm trying to explain.  LIME is also called the model-agnostic approach due to its functional independence,i.e.,  zero-intervention with the functionality of model internals. This feature makes LIME the only method to explain the cnn$_{explain}$ without seeing its internal functions.

\begin{algorithm} \label{alg:lime}
  \SetAlgoLined

\KwData{ Image $x$, number of features $n$,$cnn_{explain}$ model }
 \KwResult{ Essential segments of class $y$, segments$_{y} $}
 
segments $s$ =  Cluster($x$) \\
combinations $c$ = $\frac{S!}{r!\left(S-r\right)!}$, where r = S-1, S-2 \\
\If{$!(s \in c)$}
{
$s \leftarrow$ average as $s\prime$ for $s\in c$
}
$Px  = 150 (c) \cdot s$  
\Comment{perturbed images. }    \\ 
 
 $P$ = cnn$_{explain}$.predict($Px$)
 \Comment{predictions. } 
 
    \For {$p_{x} \in Px$ }{ 
          $ d=\frac{p_{x}\cdot {x}}{|p_{x}| |x|}  $ \\
             $D \leftarrow D \cup d$ \\
       return  $D$  \\
        \Comment{return distances $D$. } 
       }
       $d_min \leftarrow 0$ \\
       $d_max \leftarrow 1$ \\
       
  \For {$d \in D$ }{ 
   $W \leftarrow d$ \\
               $LM \leftarrow createLinear()$ 
               \Comment{create interpretable linear model}\\
               $y \leftarrow LM$.predict $(PX.W)$ \Comment{predictions of desired class $c$ using $LM$}.\\
               segments$_{y} =$ coefficients$(n,LM)$
               \Comment{ $n$ most significant coefficients of LM}
     }

 \caption{LIME.} 
\end{algorithm}

Algorithm \ref{alg:lime} presents the steps required for the explainability by the LIME. The first step segments the input image to be explained with the help of a quick-shift segmentation algorithm. The algorithm computes the segments, also known as super-pixels or features in the image. Line 2 in Algorithm \ref{alg:lime} creates the possible combinations of the segments. This involves including or excluding these segments by the factorial product of the total segments S! and segments r! at each combination. The inclusion of segments in each combination provides better differentiation later in the model. Line 3-6 in the algorithm creates 150 perturbed images based on the segment's activations in 150 random combinations. If a segment is excluded in the combination, an average of the other segments is added to keep the image size excluded from meaningful information.

Line 6 shows the Covid-19/non-Covid-19 class predictions with the $cnn_{explain}$  on the 150 perturbed images.  This returns a class core for each perturbed image for each class.  Line 8-13 starts a loo over the set of perturbed images which computes a cosine distance towards the original image. This distance measures the similarity or differences of the permuted image with the original image containing all the feature segments in it. Lines 14 and 15 assign these distances in a range between 0 and 1 to interpret the values as weights of importance for each perturbation image.  Step 16 to 19 creates an easily interpretable linear model in which outputs are the predictions towards the covid-19/non-Covid-19 class. These predictions are based on the perturbed images with their respective importance of weights.  Lines 20-21 assign the most significant features for a class in an image, i.e.,  the number of segments with more significant coefficients in the linear model. The larger the coefficient, the larger the impact on the class prediction. For a visual representation, we will plot only the essential segments regarding the Covid-19 or non-Covid-19 class, and mask the rest of the image.

\section{Explainability of cnn$_{explain}$ for Covid-19 Detection using Dry Cough }
This section presents the performance evaluation and analysis of XAI methods for Covid-19 detection using dry cough audio. It describes hardware/software, dataset, preprocessing, model, and hyperparameters.

\subsection{Experiment Setup} \label{experiment}

We have used Linux OS, intel Xeon CPU processor at 2.00GHz with 2 CPU cores,  13GB of RAM, and 24 gigabytes of VRAM from a Tesla K80 graphic card. To implement the CNN model and XAI methods, we have considered libraries including Keras, a deep learning library built on top of TensorFlow, an open-source library for machine learning tasks;tf-Keras-vis, a deep learning visualization toolkit for Keras and LIME for interpretable model-agnostic explanations. Further, we have also considered  Librosa \cite{Librosa2021} which is a package for music and audio analysis.

We have used a public dataset \cite{Kaggle2021} that consists of  160 Covid-19 positive audios. The dataset includes 19 positive classes and  21 random negative classes. The reduction of negative classes at this stage is to address the issue of imbalanced classes, later used for training and predictions. 

\subsection{Pre-processing/Augmentation}
The 40 original audios are augmented using time-stretching the audio from 30\% to 190\% of the original audio. It is further combined with other techniques including noise aggregation which adds random samples of values dispersed at regular intervals with a mean of 0 and a standard deviation of 1. Moreover, a time-shifting method that moves the sound to the beginning/end of the audio along the time axis with an interval of 15\% of the sample rate of 44,100 Hz is also used. These augmentation techniques add 2720 new audios from 40 samples,  resulting in a total of 2760 audios.

Further to augmentation, we have also converted the audios to a spectrogram using a sample rate of 44100 hertz and computing the magnitude spectrogram, and mapping it to a Mel scale. The conversion uses 1024 samples per fast Fourier transform window, 512 overlapping samples between successive frames to avoid information loss, and 128  samples of Mels. Mels represent bins of the spectrogram to its height.  Finally, the data in the spectrogram is distributed by exchanging black and white colors in the image and flipping them from top to bottom of the spectrogram.

\subsection{Model Parameters and Hyperparameters}

The $cnn_{explain}$ model consists of 3 convolutional layers and two fully connected layers. The input layer takes the spectrogram as $128 \times 820$ pixels along with three channels representing red, green, and blue color space. The convolutional layer is 2D and contains 16 units of filters. The output layer uses the max-pooling process for complexity reduction resulting in an image shape of $64 \times 410$ pixels.  We have considered a dropout of 20\% for the output layer for regularizing the $cnn_{explain}$ to prevent overfitting. The second convolutional layer consists of 32 filters. This layer also has a max-pooling process that reduces the image size to $23 \times 205$ pixels, and a dropout of 20\% of connections. Finally, the third convolutional layer consists of 64 filters followed by a max-pooling with $16 \times 102$ with a dropout of 20\%.

The $cnn_{explain}$ model uses a total of  6,708,450 trainable parameters. We have conducted experiments with 20 training epochs per iteration with a batch size of 128, and have trained the model using the Adam optimizer. Adam optimizer is known for its best performance for CNN. The learning rate to train the $cnn_{explain}$  is kept at 0.001. The activation function considered for all the layers is ReLu with the Softmax at the output layer.  An early stopping mechanism is used to avoid overfitting and accelerate training, if accuracy does not increases in 5 epochs.

\subsection{Performance Analysis}

As it is important to keep a balance between the positive and negative classes to avoid bias during the model training, this section presents an exploratory analysis of data. Further, we have evaluated the performance of the $cnn_{explain}$  in terms of sparse cross-entropy loss and accuracy as discussed below.

\begin{figure}[!t]
\centering
\includegraphics[width=3.5in]{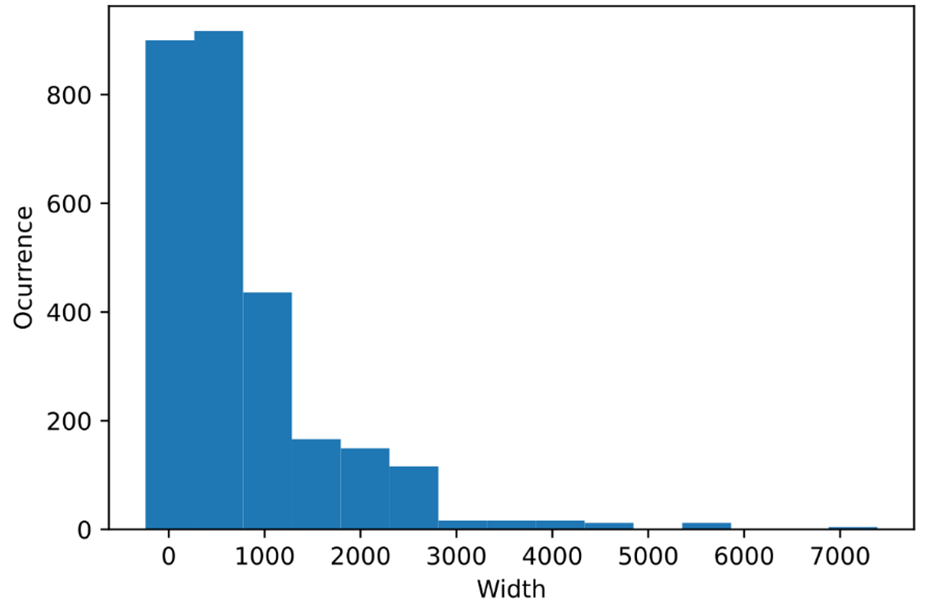}
\caption{Distributions of images widths after augmentations.}
\label{augdata}
\end{figure}

Figure \ref{augdata} presents a visualization of the image's shapes in the dataset, precisely the width of images that vary after the augmentation of the audios.  The $cnn_{explain}$ only accepts input images with the same shapes, therefore, this visual analysis provides insights for the best width  for the resizing of the images. The localization of the group of images with the same width size is between 500 and 1000 pixels followed by the group between 0 and 500. Although, these groups are close enough for resizing, there are some outliers in categories of 3,000 pixels up to 7,000 pixels. Statistically, the mean of the group is 1,014 pixels and the median 718. Thus,  all  the images are resized with a width of 820 pixels and a height of 128 pixels.

\begin{figure}[!t]
\centering
\includegraphics[width=3.5in]{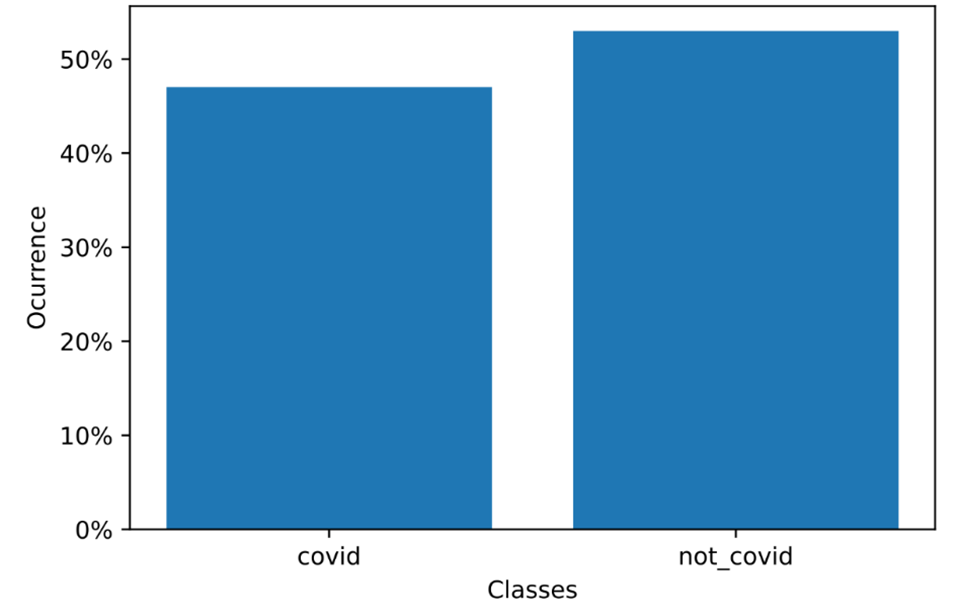}
\caption{Classes distribution in train set.}
\label{trainDist}
\end{figure}

\begin{figure}[!t]
\centering
\includegraphics[width=3.5in]{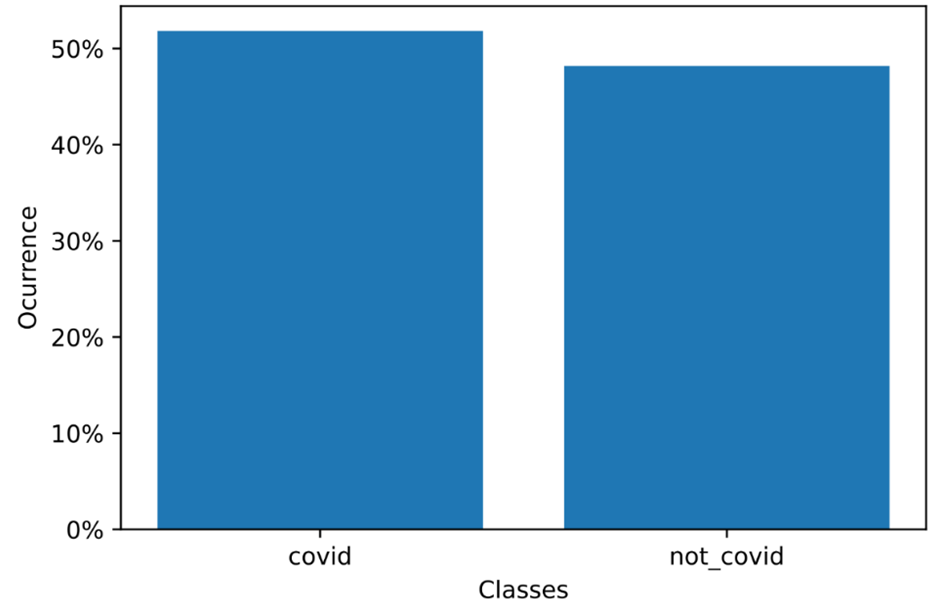}
\caption{Classes distribution in test set.}
\label{testDist}
\end{figure}

Figure \ref{trainDist} and Figure \ref{testDist} shows the distribution of Covid-19 and non-Covid-19 classes in the dataset. The expected distribution of the dataset is equal because the classes are collected in similar distributions during the data collection process. Figure \ref{trainDist} and Figure \ref{testDist}  shows the distribution of classes after a random split of data between train and test set to avoid an imbalanced dataset that can lead to inaccurate learning and predictions by the $cnn_{explain}$. Figure \ref{trainDist} and Figure \ref{testDist} show a good balance of both the positive and negative classes, thereby assuring the efficiency of the learning and testing process.

\begin{figure}[!t]
\centering
\includegraphics[width=3.5in]{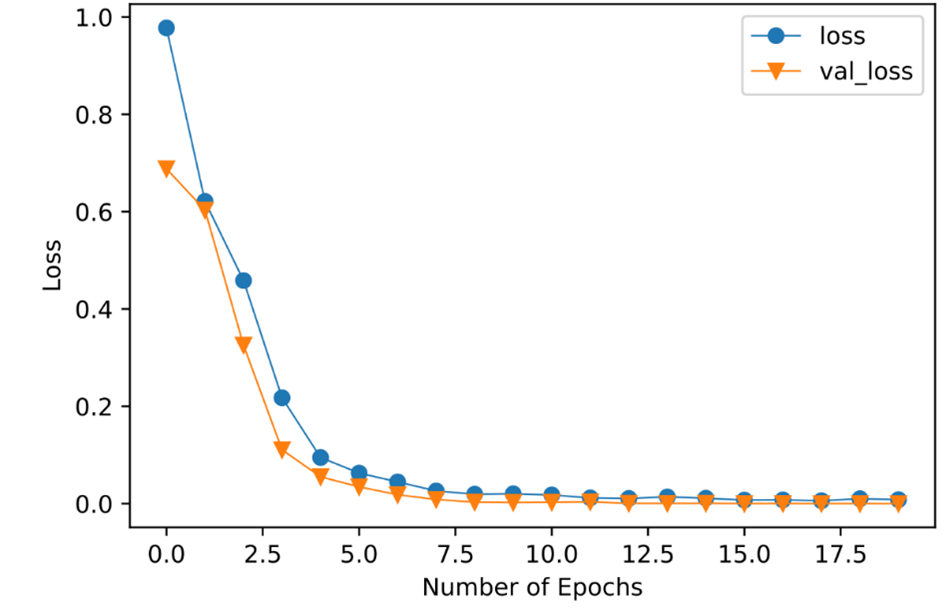}
\caption{Loss vs. number of epochs.}
\label{modelloss}
\end{figure}

\begin{figure}[!t]
\centering
\includegraphics[width=3.5in]{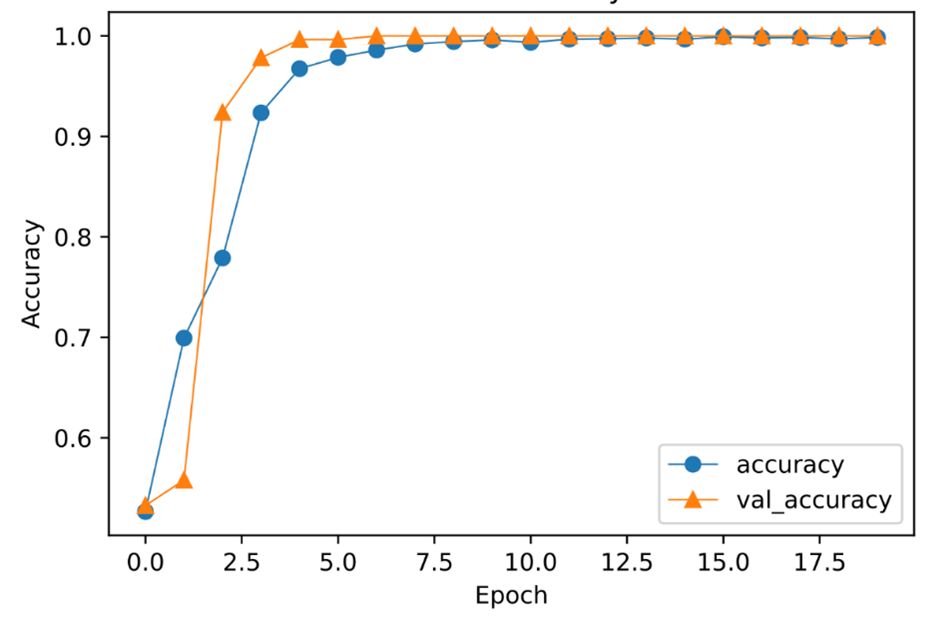}
\caption{Accuracy vs. number of epochs.}
\label{accuracymodel}
\end{figure}

To evaluate the accuracy of the model, we have changed the number of epochs from 1 to 18.  Figure \ref{modelloss} shows that initially, both the training and validation loss is higher, however, it drops significantly after the first few epochs, indicating an optimal weights adjustment by the network for a correct prediction. Finally, it fully converges after 8 epochs. 
Figure \ref{accuracymodel} shows the accuracy of the model for the training and validation. Initially, $cnn_{explain}$ demonstrates low accuracy for both the training and validation. However, after 8 epochs model is fully learned, thereby showing good accuracy.

\section{Explainability Results for $cnn_{explain}$}
This section presents the explainability results of the $cnn_{explain}$ model to understand its learning and predictions and internals. Specifically, we analyze four methods discussed in Section \ref{exmethod} to explain the $cnn_{explain}$.

\subsection{Features Visualization}
This section presents feature visualization in layers to explain the  $cnn_{explain}$
internals. Specifically, it reveals that the last five filters of the three convolutional layers of the $cnn_{explain}$  as optimal to learn parameters. Further, there is a visualization of the last dense layer to interpret the most significant features contributing to classification. These features distinguish the characteristics of each class for classification with minimum error.

\begin{figure}[!t]
\centering
\includegraphics[width=3in]{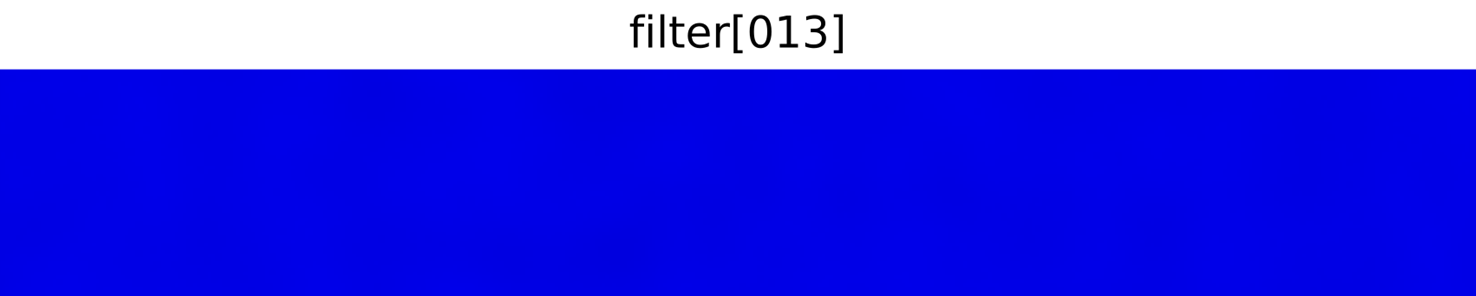}
\includegraphics[width=3in]{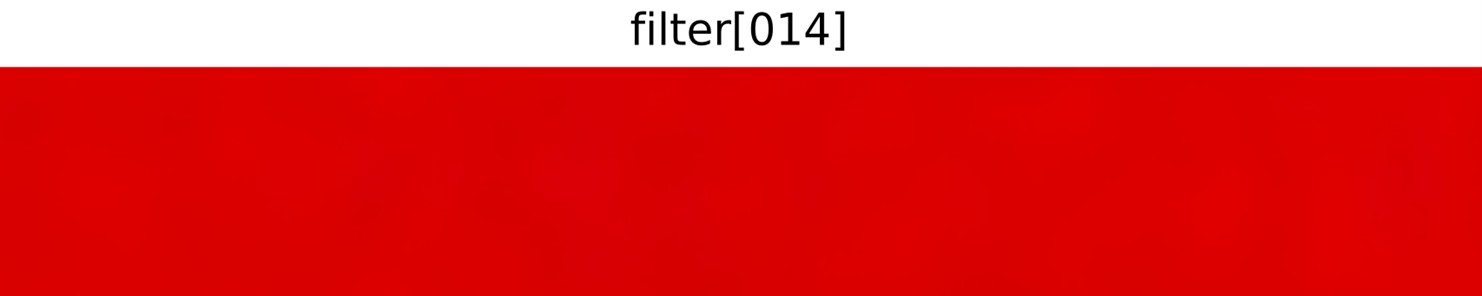}
\includegraphics[width=3in]{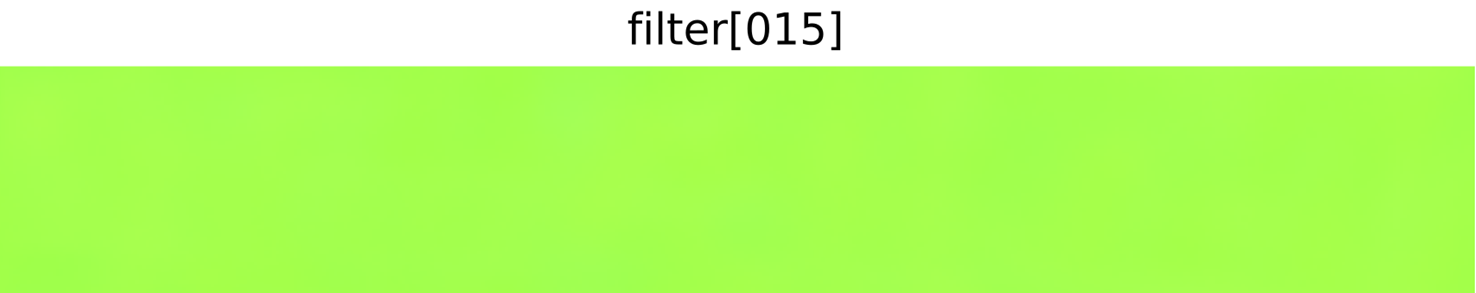}
\includegraphics[width=3in]{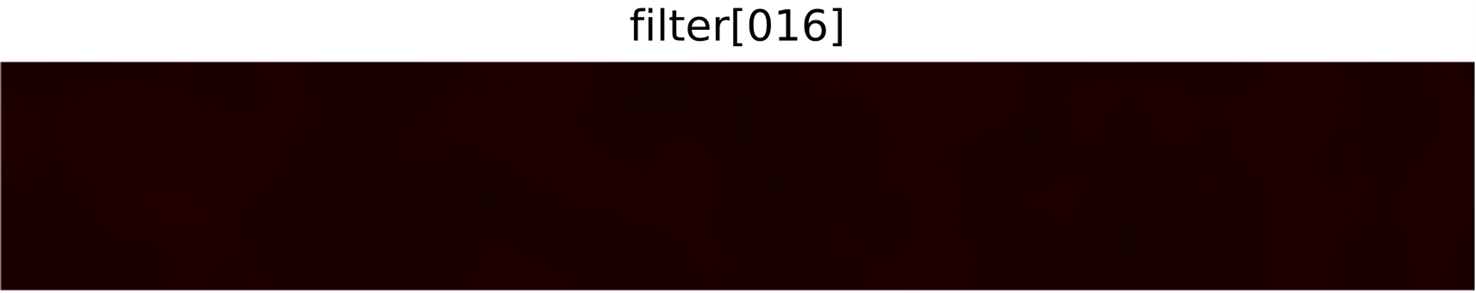}
\caption{Convolutional filters of first convolutional layer.}
\label{featurecn1}
\end{figure}

\begin{figure}[!t]
\centering
\includegraphics[width=3in]{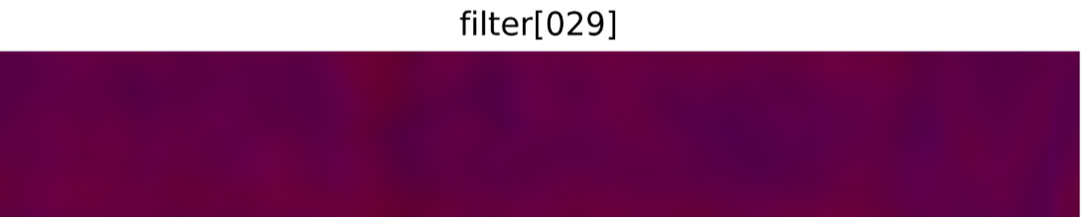}
\includegraphics[width=3in]{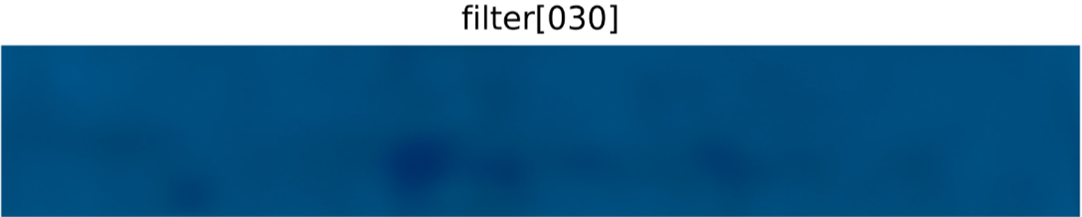}
\includegraphics[width=3in]{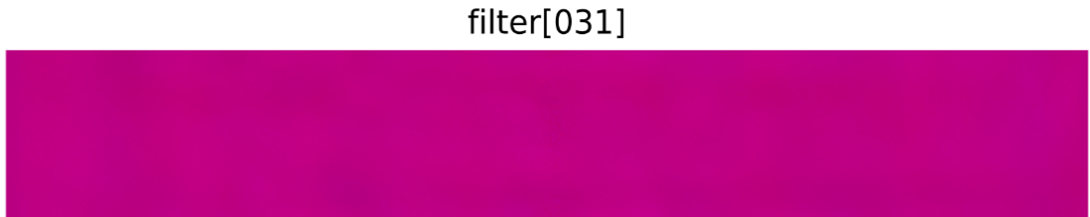}
\includegraphics[width=3in]{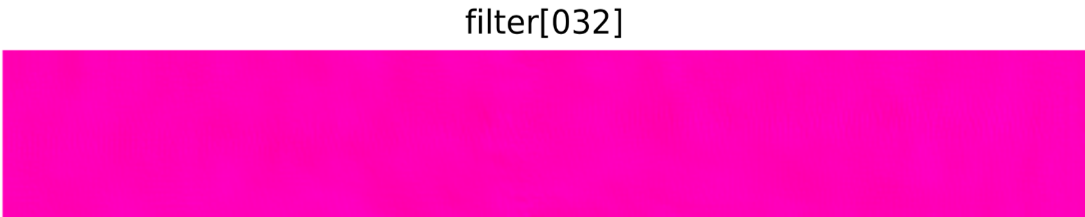}
\caption{Convolutional filters of second convolutional layer.}
\label{featurecn2}
\end{figure}

Figure \ref{featurecn1}  demonstrates the patterns learned by the first convolutional layer of the $cnn_{explain}$. This layer contains 16 units. The features learned by this layer maximize the representation of images to aid the classification of input images.  It can be noted that there is no pattern in the first convolutional layer that humans can identify more than a single color. However, these colors are a fundamental basis of the $cnn_{explain}$  to form more complex ways of understanding spectrograms in complex neural networks.
Figure \ref{featurecn2}  shows the last five filters of the $cnn_{explain}$
 in its second convolutional layer consisting of 32 units of them. Similar to the first convolutional layer, the patterns learned by the $cnn_{explain}$ are still simple not to differentiate any clear pattern in a spectrogram. However, in filter 29, the primary color learned by the filter is red, but there are spotted areas with a slightly different color. However, filter number 30 has activated the features like blue color, but a few areas of stronger blue are mainly located in the center and lower part of the channel. Thus, at this stage, the appearance of patterns beyond single colors can be observed.

\begin{figure}[!t]
\centering
\includegraphics[width=3in]{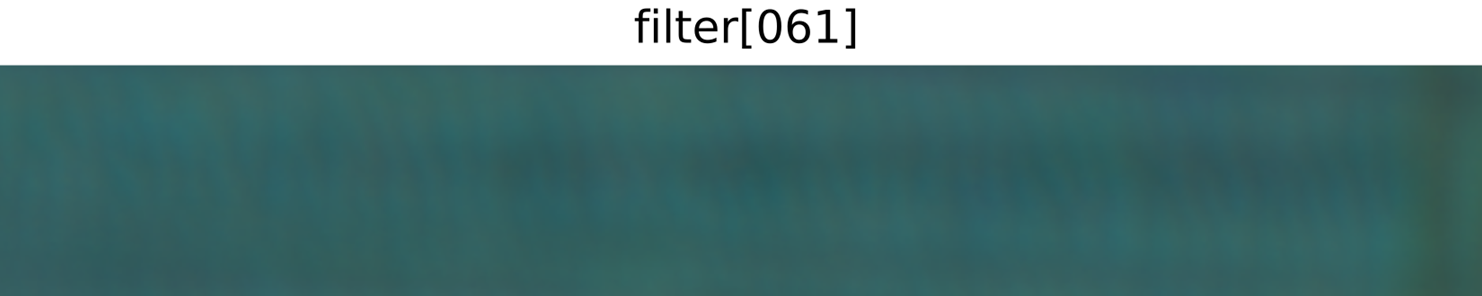}
\includegraphics[width=3in]{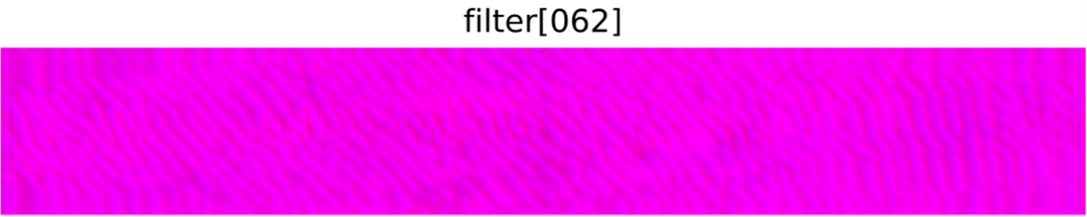}
\includegraphics[width=3in]{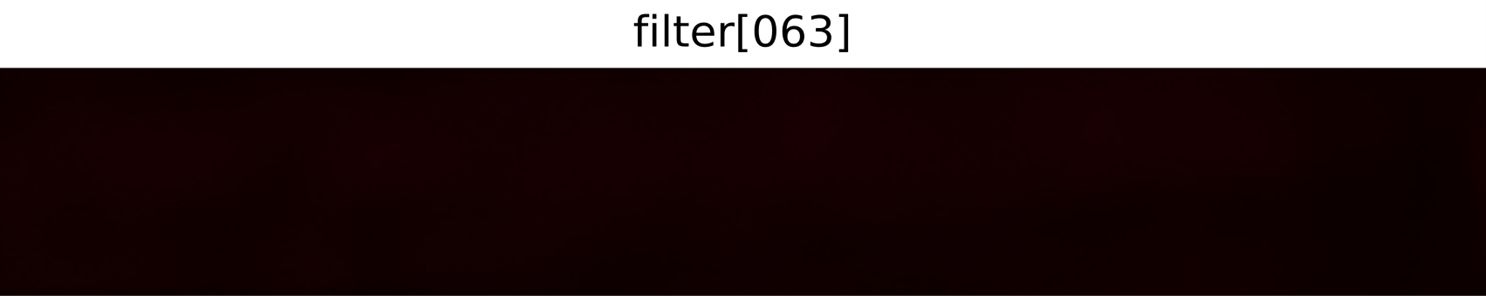}
\includegraphics[width=3in]{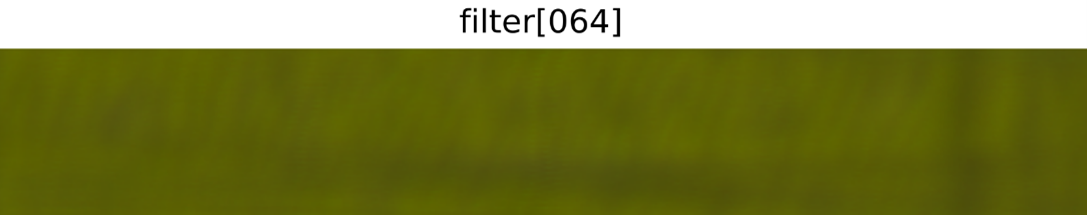}
\caption{Convolutional filters of third convolutional layer.}
\label{featurecn3}
\end{figure}

\begin{figure}[!t]
\centering
\includegraphics[width=3in]{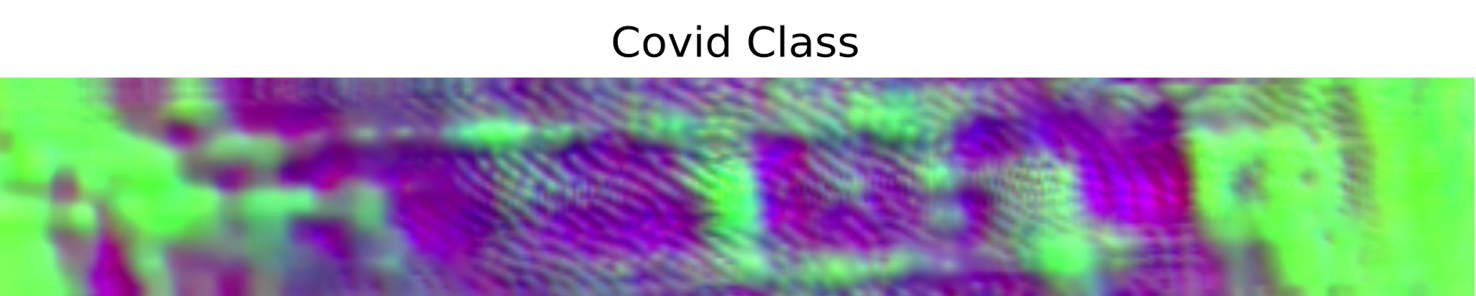}
\includegraphics[width=3in]{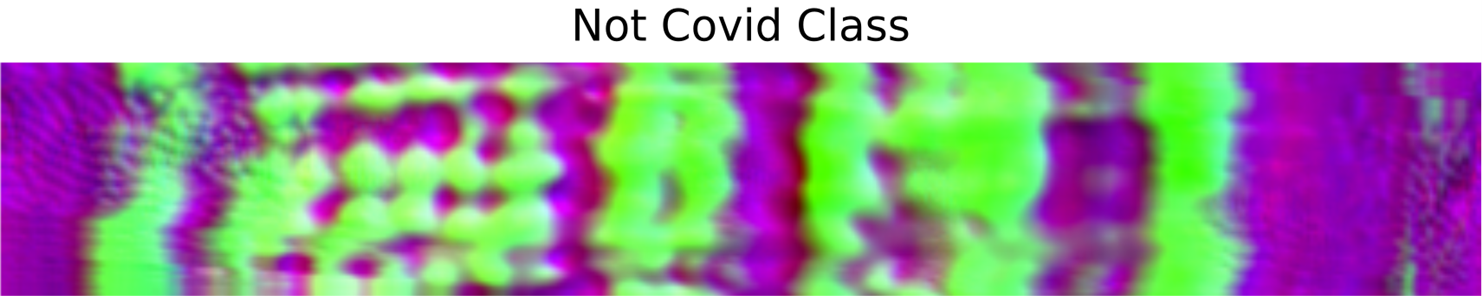}

\caption{Softmax layer.}
\label{softmx}
\end{figure}

Figure \ref{featurecn3}
 presents the filters that the CNN learned in the third and last convolutional layer with 64 channels. At this stage, most  $cnn_{explain}$
filters contain more meaningful information for class distinction. For example, filters 61, 62, and 64 show different colors and distinct shapes. The three images show some line patterns that stain the filter. The visualization shows that the filters in the  $cnn_{explain}$ 's last layer convert from spots found in previous layers to lines.
Figure \ref{softmx} shows the important input features  that $cnn_{explain}$
 can distinguish to classify each class in the last layer. These features can be seen as a combination of two colors in the spectrograms, i.e., green, and purple. The first noticeable difference between them is the stained part of the Covid-19 class, mainly located in the centre of the image. The localization of this feature is for the non-Covid-19 class with less presence in the surroundings of the spectrogram. The second difference is that the stained feature in the non-Covid-19 class appears mainly in purple,  whereas the Covid-19 class has a presence of purple and green colors. The stained feature has a correlated shape with the patterns visualized in the previous layers.

\subsection{Spectrogram Samples} \label{spctsample}
\begin{figure}[!t]
\centering
\includegraphics[width=3in]{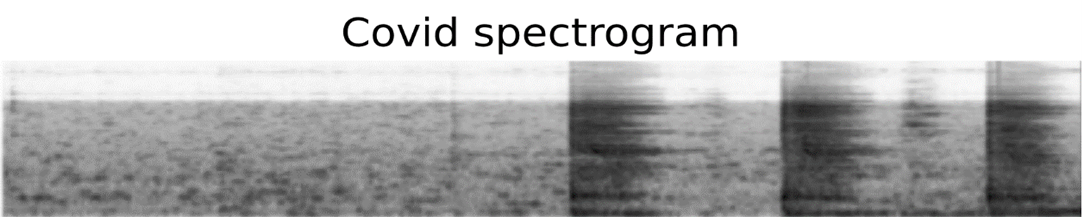}
\includegraphics[width=3in]{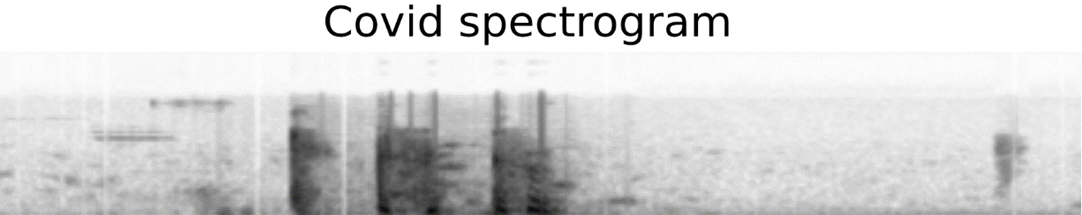}
\includegraphics[width=3in]{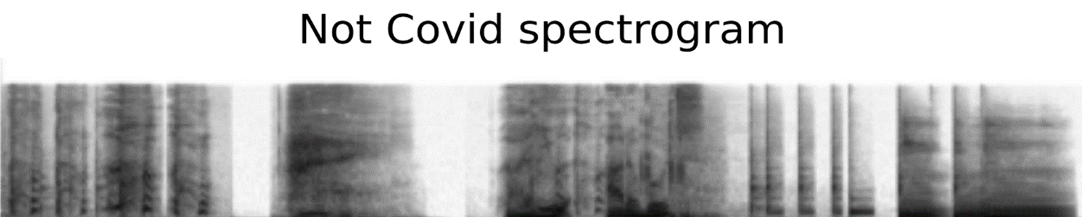}
\includegraphics[width=3in]{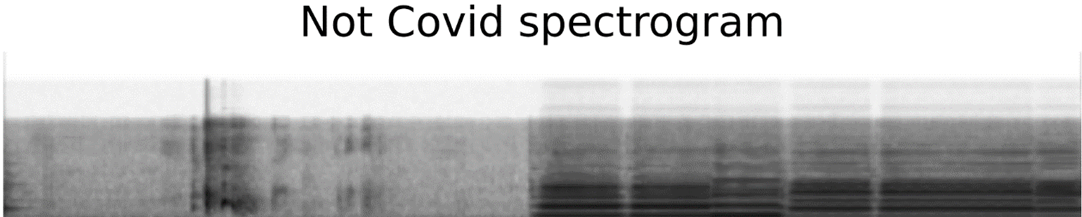}
\caption{Spectrograms samples.} \label{spectro}
\end{figure}

Figure \ref{spectro} shows four spectrogram samples that are later used for the remaining three explainable methods. For simplicity, four spectrograms are taken, with a balance of classes regarding two spectrograms representing audio of Covid-19 patients and two spectrograms representing audio of patients without Covid-19. The order of these spectrograms remains fixed throughout the following sections.

\subsection{Smooth Gradient (SmoothGrad)}

This section analyses the explainability of $cnn_{explain}$ using SmoothGrad. During the experiments, We have selected 50 smoothed images with a 50\% noise spread level. In the visualization of results, the redder parts in the maps symbolize the significant presence of the class activated in the spectrogram,  whereas the bluer ones show less. The first image represents the average gradients towards the Covid-19 class, and the second image shows the average gradients toward the non-Covid-19 class regardless of the spectrogram class.
 \begin{figure}[!t]
\centering
\includegraphics[width=3in]{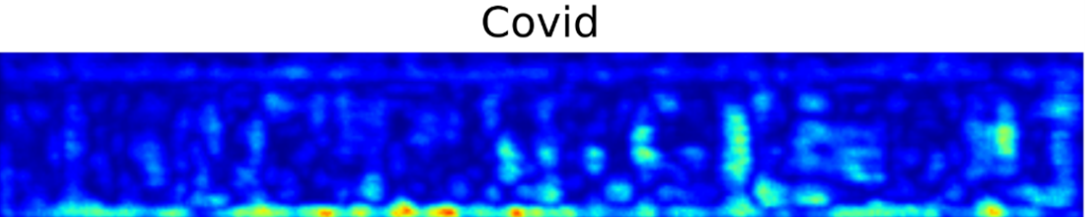}
\includegraphics[width=3in]{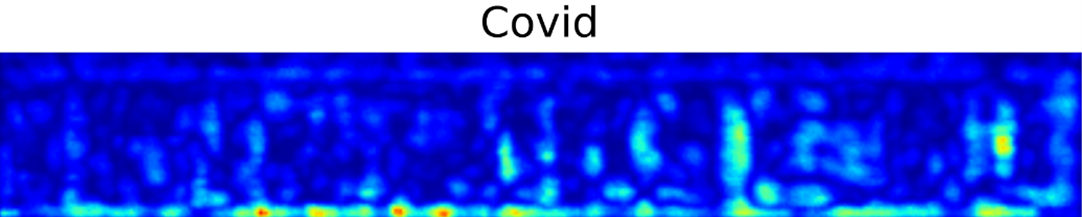}
\includegraphics[width=3in]{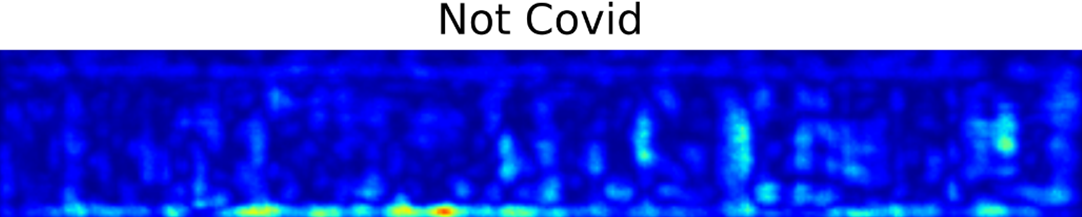}
\includegraphics[width=3in]{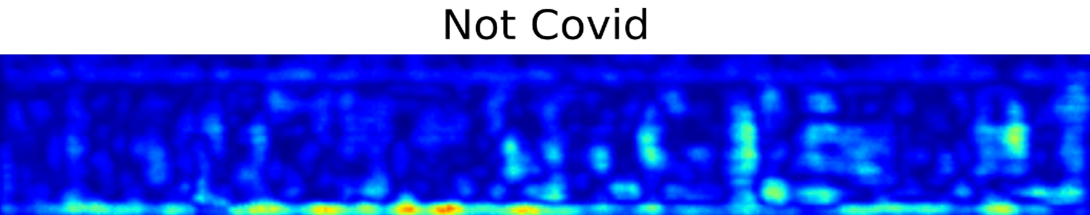}
\caption{SmoothGrad for Covid-19 class.}
\label{smthcovid}
\end{figure}
Figure \ref{smthcovid} presents the saliency maps of the SmoothGrad algorithm for the four images as given in Section \ref{spctsample} towards the Covid-19 class. The SmoothGrad method seems to underperform the distinction of the classes due to its inherent property of averaging gradients. The most likely Covid-19 area by the model is the lower part of the spectrogram in the four spectrograms. However, not all four spectrograms are in the same class. 
 \begin{figure}[!t]
\centering
\includegraphics[width=3in]{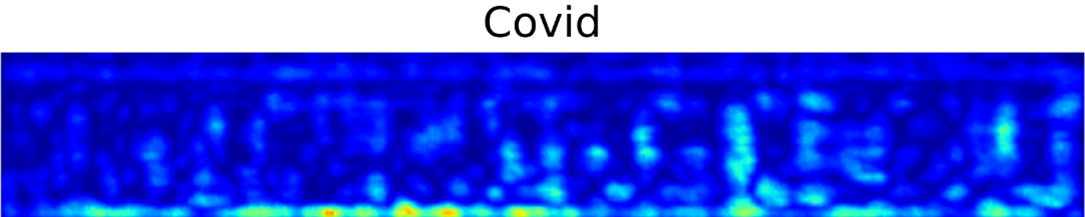}
\includegraphics[width=3in]{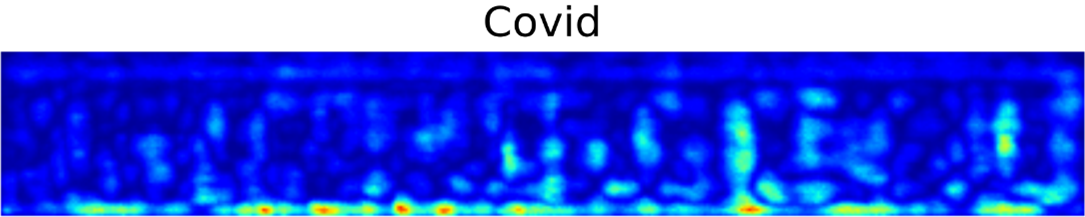}
\includegraphics[width=3in]{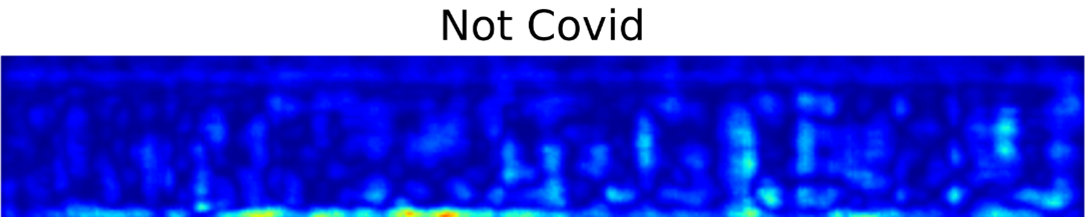}
\includegraphics[width=3in]{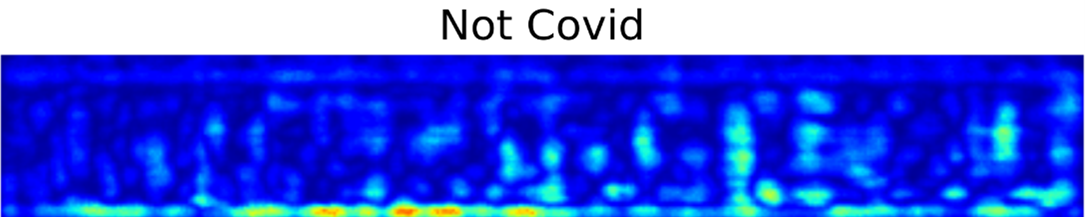}
\caption{SmoothGrad fornon-Covid-19 class.}
\label{smthnoncovid}
\end{figure}
Figure \ref{smthnoncovid} is the second representation of the SmoothGrad algorithm toward the non-Covid-19 class. Again, the algorithm seems to highlight the same areas of the four spectrograms with minor differences. Also, the highlighted areas of this  non-Covid-19 class remain the same as the Covid-19 class. Thus, it seems like an underperforming task by SmoothGrad. 

\subsection{Gradient-weighted Class Activation Mapping (Grad-CAM)}
\begin{figure}[!t]
\centering
\includegraphics[width=3in]{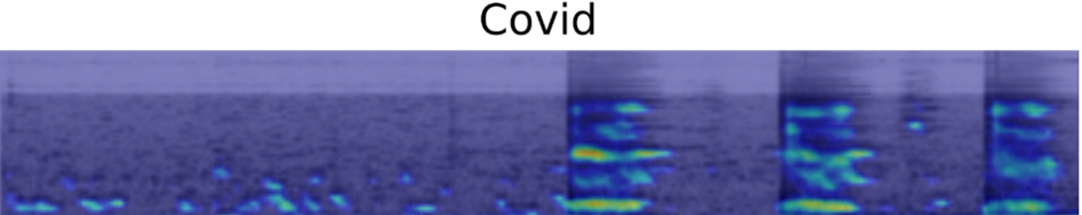}
\includegraphics[width=3in]{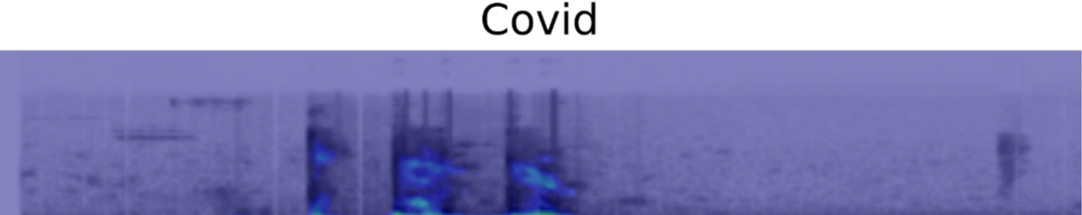}
\includegraphics[width=3in]{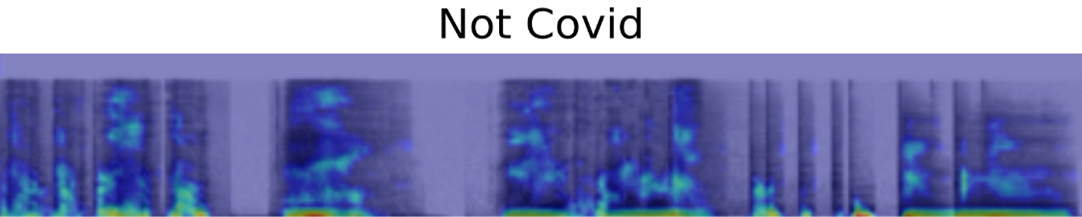}
\includegraphics[width=3in]{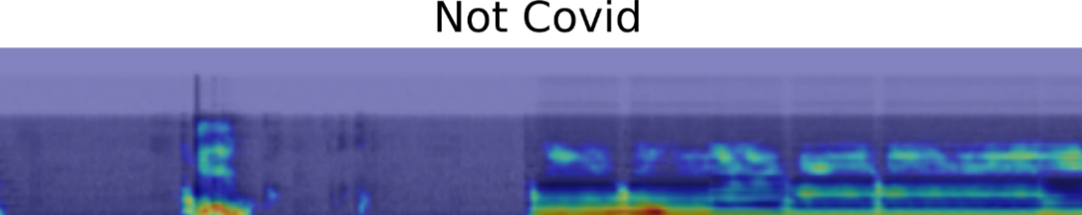}
\caption{Grad-CAM explanation for Covid-19 class.}
\label{gradcamcovid}
\end{figure}

\begin{figure}[!t]
\centering
\includegraphics[width=3in]{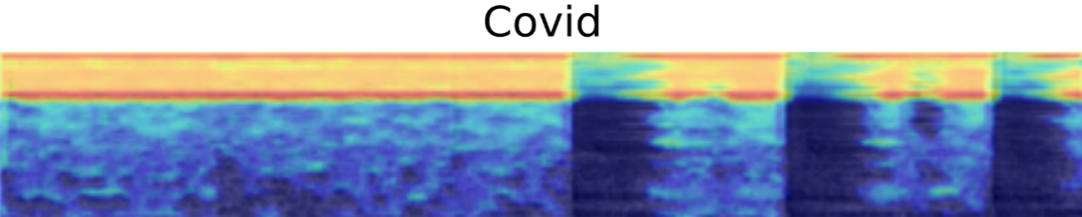}
\includegraphics[width=3in]{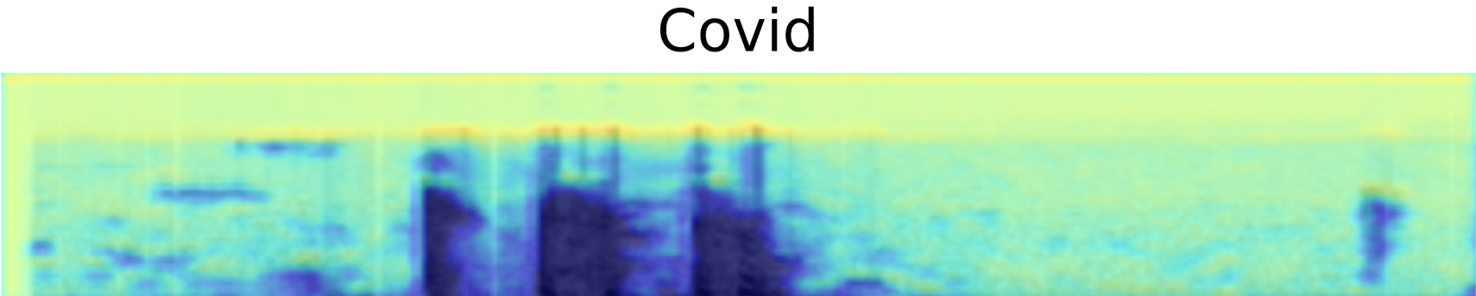}
\includegraphics[width=3in]{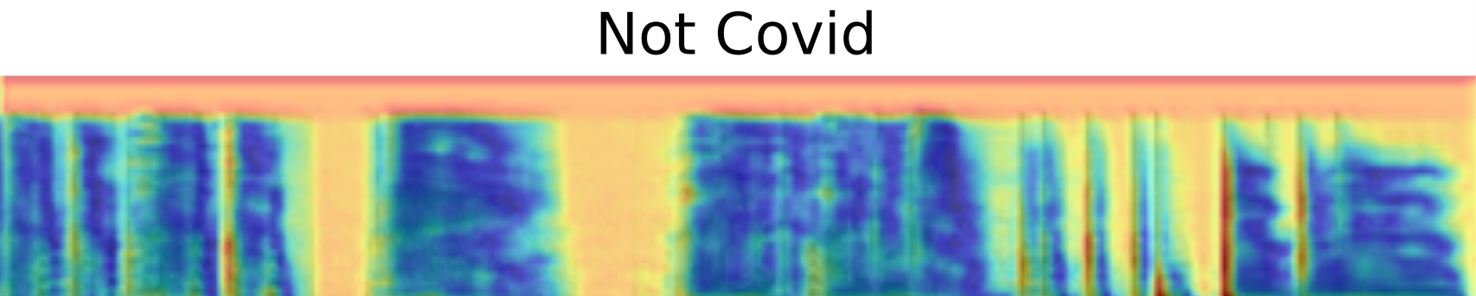}
\includegraphics[width=3in]{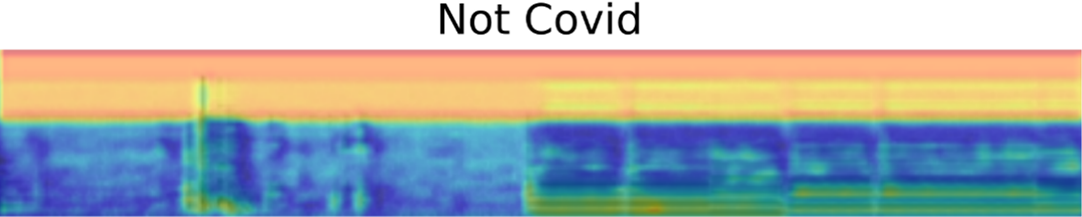}
\caption{Grad-CAM for non-Covid-19 Class.}
\label{gradnoncovid}
\end{figure}

This section presents the results obtained with the GradCAM algorithm. Similarly, as seen in the last section, the explanation of the model by the  GradCAM focuses Covid-19 and non-Covid-19 classes. 
Figure \ref{gradcamcovid} shows the four spectrograms with activations towards Covid-19, where the red colours represent the areas the Covid-19 relevant features. The first two Covid-19 spectrograms have a slightly less red presence than the last two in the area with more energy indicating non-Covid-19 class. The first two Covid-19 spectrograms contain a stained feature surrounding the area with more energy, a change in contrast to the last two spectrograms of non-Covid-19  that are coloured as blue.
Figure \ref{gradnoncovid} shows the four spectrograms with the activation of non-Covid-19 class where the red represent the areas with highlighted non-Covid features. The first two Covid-19 spectrograms contain the strongest blue areas in the image's high-energy parts, representing the less important areas for non-Covid-19. The last two, non-Covid-19 spectrograms have blue areas as well. As can be observed, the higher activation of red is dominant than the blue, representing the correct classification of non-Covid-19 case.

\subsection{Local Interpretable Model-Agnostic Explanations (LIME)}
\begin{figure}[!t]
\centering
\includegraphics[width=3in]{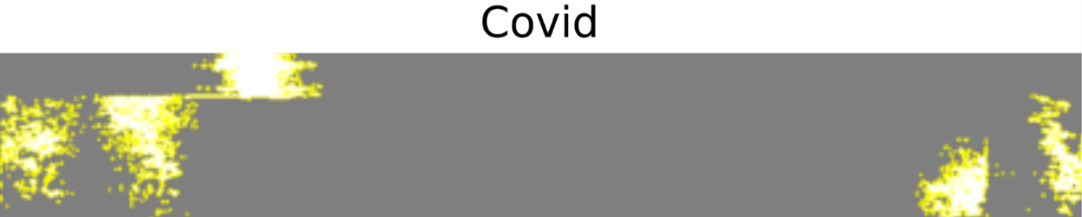}
\includegraphics[width=3in]{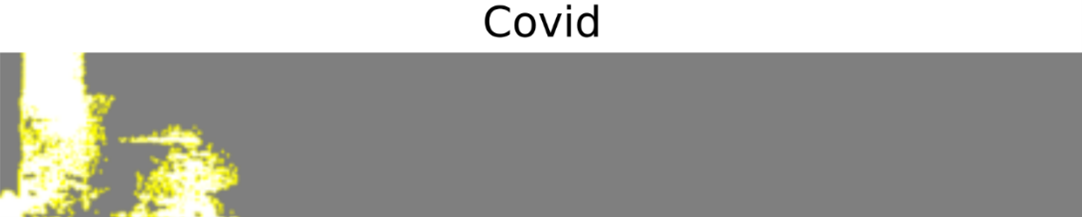}
\includegraphics[width=3in]{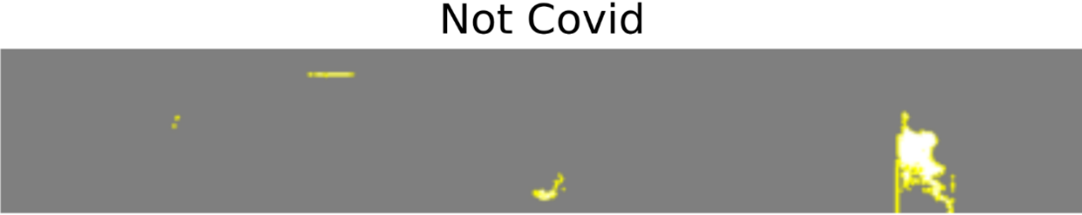}
\includegraphics[width=3in]{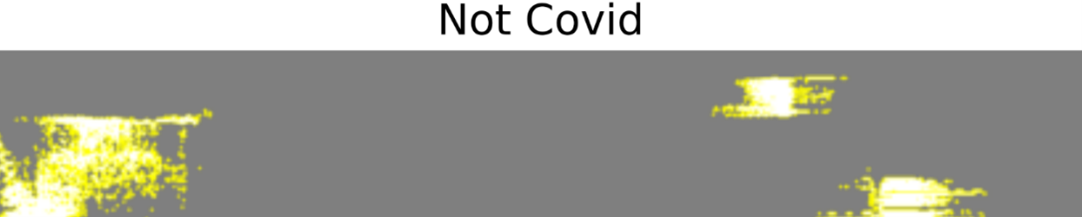}
\caption{LIME for covid-19 class.}
\label{limecovid}
\end{figure}

\begin{figure}[!t]
\centering
\includegraphics[width=3in]{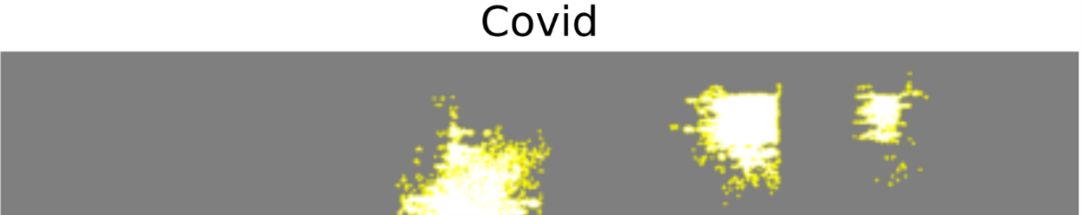}
\includegraphics[width=3in]{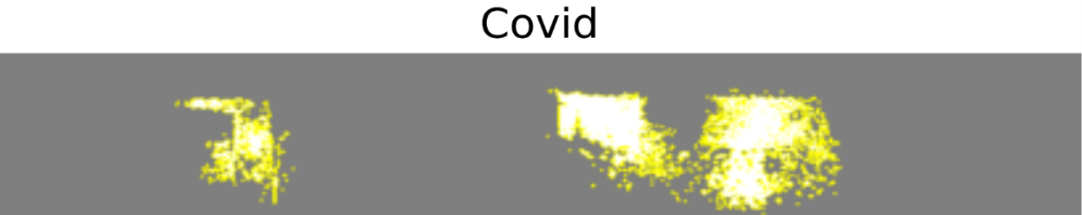}
\includegraphics[width=3in]{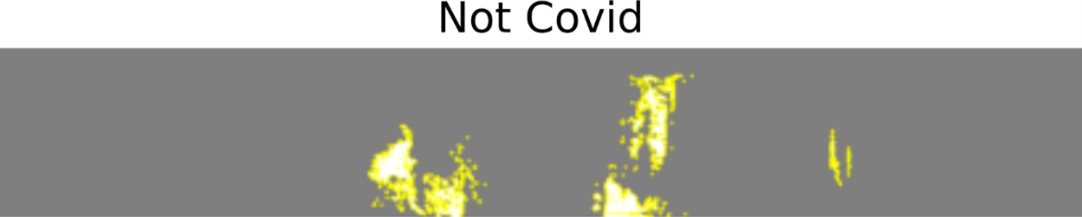}
\includegraphics[width=3in]{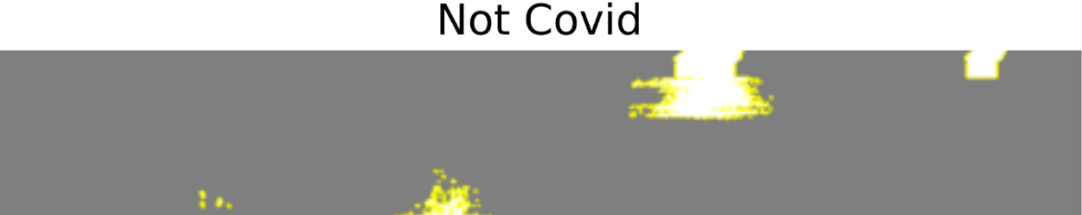}
\caption{LIME for non-Covid-19 class.}
\label{limenocovid}
\end{figure}

Similar to other explainability methods, the results with the LIME algorithm remain the same. Figure \ref{limecovid} represents the activation of the Covid-19 class and Figure \ref{limenocovid} of non-Covid class for the four spectrograms. 
Figure \ref{limecovid}
shows the four spectrogram's output from the LIME algorithm for the Covid-19 class. This approach highlights stained segments seen in the first image before and after a high-energy sound, and in the second image at the beginning of the spectrogram.It can be noted that for the non-Covid spectrograms, the areas of Covid-19 relevant features appear to have higher frequencies with fewer frequency segments without clear signs distinction. 

Figure \ref{limenocovid} shows the results of the LIME algorithm in the four spectrograms towards the non-Covid-19 class. The algorithm highlights segments with relatively low stained areas. In case of non-Covid-19 spectrograms, the algorithm highlights segments that contain waveforms with more energy of sound as seen in the third image. The fourth image also shows areas at higher frequencies.

\section{Conclusion}

Most of the deep learning approaches considered for medical decisions do not come as 'out-of-the-box', thereby limiting their applicability for domains such as medical or malware decisions, where wrong decisions can lead to harmful consequences. 
In the context of Covid19 detection, full comprehensibility of the deep
learning approach is hard to achieve, although the facilitated
predictions are useful to the practitioner.   It is crucial to validate predictions by providing interpretations of the model and the contribution of relevant features to classification.
XAI can address these challenges by producing more explainable models. It helps humans to interpret, trust and understand the emerging generation of machine learning models.  Our work presents the explainabiility of the CNN model called $cnn_{explain}$ using four XAI techniques such as visualization, SmoothGrad, Grad-CAM, and LIME. These approaches explain the classification of the  $cnn_{explain}$ model for Covid-19 and non-Covid-19. The interpretability provided by these XAI methods contributes to more reliable decisions by providing insights into an input space's relevance and analysis of intermediate relevance filtres at  $cnn_{explain}$ 's hidden layers. The results by all the XAI methods have yielded encouraging results from our effort in the explanation of CNN, i.e., $cnn_{explain}$ in terms of its processing, network representation,  filtre level explanation, and classification.

Initially, we  have evaluated the performance of $cnn_{explain}$ with a training accuracy of 98.8\% after a few epochs for Covid-19 and non-Covid-19. This ensures that $cnn_{explain}$ model fulfills the requirements of good accuracy for Covid-19 detection, thereby making the model a good candidate for the explanation.
 
The visualization of convolutional filters clearly shows the increasing complexity of patterns learned by the   $cnn_{explain}$ that allows the correct classification of the classes. Also, the dense layer shows the features that most likely represent a class by the $cnn_{explain}$.  The results show that there is a stained feature that is clearly present in both classes that has a positive correlation with the shapes found in the filters. However, the presence of features is greater in the positive class than in the negative class. Further, the presence of that feature is more prominent for the purple color in the positive class, rather than the green in the negative.  

The sensitive maps from the SmoothGrad and Grad-CAM algorithm highlight important features of a sample as does the segments retrieved from the LIME algorithm. SmoothGrad aims to eliminate noise from the sensitive maps, however, it underperforms in its application on spectrograms for the use case, due to its sensitive map showing little distinction for the interpretation of classes. Finally, the remaining two algorithms show some areas that are highlighted for the positive Covid-19 class with stained segments, mainly beside the sounds with high energy. The LIME algorithms seem to highlight areas with waveform in high-energy areas for the negative non-Covid-19 class. Unlike the Grad-CAM algorithm that seems to relate to higher frequency areas for the negative class.

Furthermore, results of the LIME show it as a suitable explanation model for neural network architectures such as convolutional neural networks, or for types of classification problems (e.g., Covid-19 and non-Covid-19). More generally, interpretability by these selected models can contribute to the design of accurate and efficient classifiers, not only by analyzing and leveraging the features' relevance but also through the
analysis of intermediate relevance features “at classifier 's filtres within hidden layers”. The results from this research have yielded encouraging results from our efforts in the explanation of convolutional neural network processing with a filter-level explanation.

\section*{Acknowledgments}
This work is supported by the Atlantic Technological University (ATU), Donegal Campus, Ireland.


\newpage

\begin{IEEEbiography}[{\includegraphics[width=1in,height=1.25in,clip,keepaspectratio]{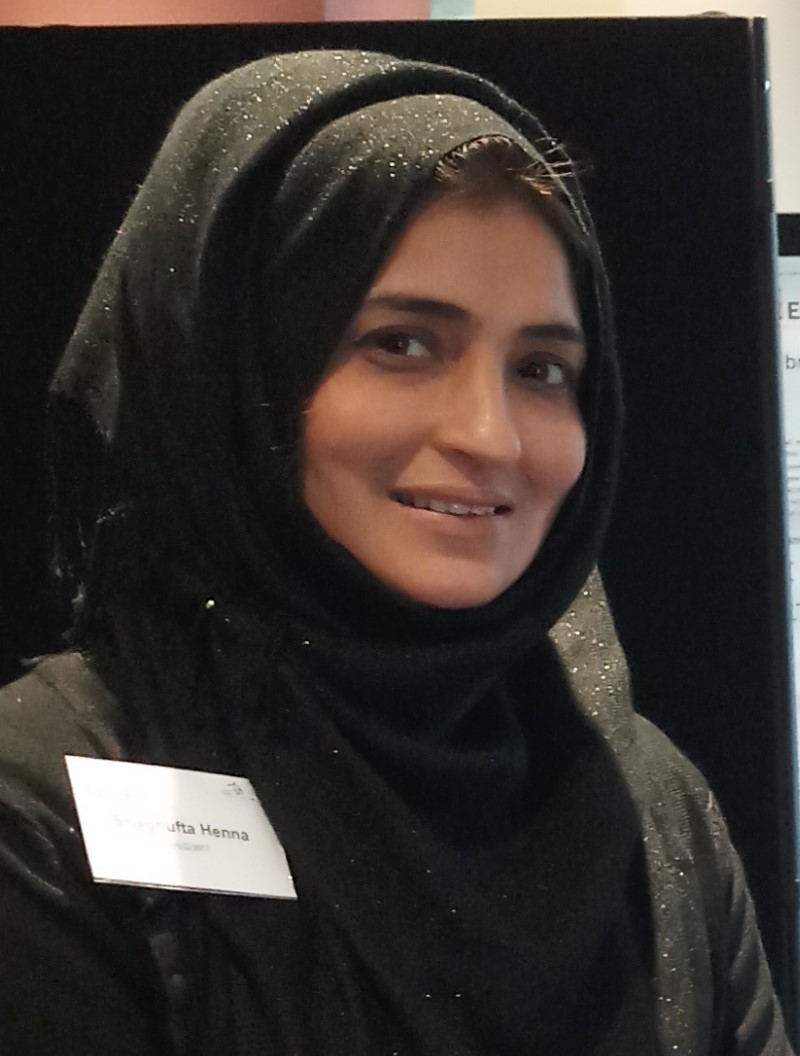}}]{Shagufta Henna} is lecturer with the Atlantic Technological University, Donegal, Ireland. She was a post-doctoral researcher with the  Waterford institute of technology, Waterford, Ireland from 2018 to 2019. She received her doctoral degree in Computer Science from the University of Leicester, UK in 2013. She is an Associate Editor for IEEE Access,EURASIP Journal on Wireless Communications and Networking, IEEE Future Directions, and Human-centric Computing and Information Sciences, Springer. She is senior member of IEEE. Her current research
interests include edge computing, self-supervised learning, attention in deep learning, representation learning, deep reinforcement learning, explainable AI, and AI-driven network optimizations.
\end{IEEEbiography}

\vspace{11pt}
\begin{IEEEbiographynophoto}{
Juan Miguel Lopez Alcaraz
}
 is a Master stundent of Big Data Analytics and Aritificial Intelligence. His research interests include machine learning, explainable AI, and big data analytics.
\end{IEEEbiographynophoto}

\vfill

\end{document}